\documentclass{article}

\usepackage{microtype}
\usepackage{graphicx}
\usepackage{booktabs} %

\usepackage{hyperref}

\usepackage[accepted]{icml2019}

\usepackage{soul}
\usepackage[dvipsnames]{xcolor}
\usepackage{dsfont}

\usepackage{amssymb}
\usepackage{amsmath}

\usepackage{subcaption}

\DeclareMathOperator*{\argmax}{arg\!\max}

\icmltitlerunning{Dynamic Weights in Multi-Objective Deep Reinforcement Learning}
 
\begin{document}
 
\twocolumn[
	\icmltitle{Dynamic Weights in Multi-Objective Deep Reinforcement Learning}
\icmlsetsymbol{equal}{*}

\begin{icmlauthorlist}
	\icmlauthor{Axel Abels}{ulb,vub}
	\icmlauthor{Diederik M. Roijers}{vua}
	\icmlauthor{Tom Lenaerts}{ulb,vub}
	\icmlauthor{Ann Nowé}{vub}
	\icmlauthor{Denis Steckelmacher}{vub}
	\end{icmlauthorlist}
	
	\icmlaffiliation{ulb}{Machine Learning Group, Université Libre de Bruxelles, Brussels, Belgium}
	\icmlaffiliation{vub}{Artificial Intelligence Lab, Vrije Universiteit Brussel, Brussels, Belgium}
	\icmlaffiliation{vua}{Computational Intelligence, Vrije Universiteit Amsterdam, Amsterdam, the Netherlands}
\icmlcorrespondingauthor{Axel Abels}{aabels@ulb.ac.be}
\icmlkeywords{Multi-Objective Reinforcement Learning, Deep Reinforcement Learning}

\vskip 0.3in
]

\printAffiliationsAndNotice{} %

\begin{abstract}
	Many real world decision problems are characterized by multiple
conflicting objectives which must be balanced based on their relative
importance. In the dynamic weights setting the relative importance
changes over time and specialized algorithms that deal with such change,
such as the tabular Reinforcement Learning (RL) algorithm by \citeauthor{Natarajan:2005:DPM:1102351.1102427} (\citeyear{Natarajan:2005:DPM:1102351.1102427}), are required. However, this earlier work is not feasible for RL settings that necessitate the use of function approximators. We generalize across weight changes and high-dimensional inputs by proposing a multi-objective Q-network whose outputs are conditioned on the relative importance of objectives, and introduce Diverse Experience Replay (DER) to counter the inherent non-stationarity of the dynamic weights setting. We perform an extensive experimental evaluation and compare our methods to adapted algorithms from Deep Multi-Task/Multi-Objective RL and show that our proposed network in combination with DER dominates these adapted
algorithms across weight change scenarios and problem domains.

\end{abstract}

\noindent

\section{Introduction}

In \emph{reinforcement learning (RL)} \cite{sutton1998reinforcement}, an agent learns to behave in an %
 unknown environment based on the rewards it receives. In single objective RL these rewards are scalar. However, most real-life problems are more naturally expressed with multiple objectives. For example, autonomous drivers need to minimize travel time and fuel consumption, while maximizing safety \cite{DBLP:journals/corr/XiongWZL16}.

When user utility in a multi-objective problem is defined as a linear scalarization with weights per objective that are known in advance and fixed throughout learning and execution, the problem can be solved via single-objective RL. However, in many cases the weights can not be determined in advance \cite{roijers2017multi} or linear scalarization does not apply because the user utility cannot be expressed with a linear function \cite{JMLR:v15:vanmoffaert14a}. In this paper, we focus on the setting where the weights are linear, but not fixed. Specifically, the parameters of the scalarization function change over time. For example, if fuel costs increase, a shorter travel time could no longer be worth the increased fuel consumption. This is called the \emph{dynamic weights} setting \cite{Natarajan:2005:DPM:1102351.1102427}. %

Many RL problems necessitate learning from raw input, e.g., images captured by cameras mounted on a car.
Recently, Deep RL \cite{DBLP:journals/corr/MnihKSGAWR13} has enabled RL to be applied to problems where the input consists of %
images.
However, most Deep RL research focuses on %
single-objective problems. %

In this paper, we study the possibilities of Deep RL %
in the dynamic weights setting %
and show how transfer learning techniques can be leveraged to increase the learning speed by exploiting information from past policies. For tabular RL, these principles have previously been applied to, e.g., Buridan's ass problem \cite{Natarajan:2005:DPM:1102351.1102427}. However, because of its small and discrete state space %
this problem is not representative of complex real-world problems which often have vast or even continuous state spaces. In such complex problems, tabular RL is not feasible.
 
To tackle high-dimensional problems, we show that algorithms from related settings can be adapted to the dynamic weights settings but are inadequate. We therefore propose the \emph{conditioned network (CN)}, in which a Q-Network is augmented to output weight-dependent multi-objective Q-value-vectors.
To efficiently train this network, we propose an update rule specific to the dynamic weights setting. We further propose \emph{Diverse Experience Replay (DER)}, to improve sample-efficiency and reduce replay buffer bias. %

To benchmark the quality of our algorithms, we propose the first non-trivial high-dimensional multi-objective benchmark problem:  \emph{Minecart}. From raw visual input, an agent in Minecart must learn to %
adapt to the day's valuation of different resources to efficiently mine them while minimizing fuel consumption.
We test the performance of our algorithms on two weight change scenarios and find that, while methods from related settings can be adapted to the dynamic weights setting, only our proposed CN can both quickly adapt to sparse abrupt weight changes and also converge to optimal policies when weight changes occur regularly. Furthermore, by maintaining a set of diverse trajectories, DER improves the performance of all tested algorithms.

\section{Background}

This section defines Markov Decision Processes and Q-Learning, then briefly reviews the Deep RL literature and Multi-objective RL.%

\subsection{Markov Decision Process}
In RL, agents learn how to act in an environment in order to maximize their cumulative reward. A popular model for such problems is Markov Decision Processes (MDP), defined by a set of states $S$, a set of actions $A$, a transition function $T$ which maps the state $s_t$ and action $a_t$ to a probability over all possible next states $s_{t+1}$, and a reward function $R$ which maps each state $s \in S$ and action taken in it to an expected immediate reward $r_t = R(s_t, a_t)$. Under the standard assumption that future rewards are discounted by a factor $\gamma \in [0,1]$, the goal of the agent is to find a policy $\pi^*(a | s)$ that maximizes the expected cumulative reward of the agent, i.e., its \emph{return},
$	g_T = \sum_{t=1}^T \gamma^{t-1} r_t$. We define a \emph{trajectory} $\tau$ as a sequence of transitions from some state $s_i$ to a state $s_{j+1}$; $\tau = [(s_i,a_i, r_i,s_{i+1}),...,(s_j,a_j, r_j,s_{j+1})] $.

The value function $V^{\pi} : S \rightarrow \mathds{R}$ of a policy $\pi$ maps a state to the expected return obtained from that state, when $\pi$  is followed, i.e., $V(s) = E_{\pi} [\sum_{t=1}^{\infty} \gamma^{t-1} r_t | s_1 = s]$. Correspondingly, the Q-function $Q^{\pi} : S \times A \rightarrow \mathds{R}$ maps a state-action pair to the expected return obtained from that state when the action is executed, and then $\pi$ is followed from the next state onwards.
The value function $V^*$ and Q-function $Q^*$ that correspond to the optimal policy $\pi^*$ are the \emph{optimal} value functions. The optimal policy $\pi^*$ can be computed from the optimal $Q^*$ function;
$\pi^*(a | s) = \mathds{1} [a = \argmax_{a'} Q^*(s, a')]$,
i.e., the agent executes, at every time-step, the action whose Q-value in the current state is maximal. This is the \emph{greedy policy} w.r.t.\ $Q^*$. %
The \emph{stateless value} for a policy $\pi$ is defined as
${V}^{\pi} = \sum_{s\in S} \mu(s) { V}^{\pi}(s)$, with $\mu(s)$ the probability distribution over initial states.

\textbf{Q-Learning} \;\;
Q-learning \cite{Watkins:1989} is a reinforcement learning algorithm that allows an agent to learn $Q^*$ for any (finite) MDP based on interactions with the environment. %
At every time-step, the agent observes the state $s_t$, executes a random action $a_t$ with probability $\varepsilon$ and $a_t \sim \pi(s_t)$ otherwise, receives a reward $r_t$, then observes the next state $s_{t+1}$. Based on this $(s_t, a_t, r_t, s_{t+1})$ \emph{experience tuple}, the agent updates its estimate of $Q^*$ at iteration $k$:
$Q_{k+1}(s_t, a_t) = Q_k(s_t, a_t) + \alpha \delta_k$, where $\delta_k = r_t + \gamma \max_{a'} Q_k(s_{t+1}, a') - Q_k(s_t, a_t)$,
with $\alpha > 0$ a small learning rate. Q-learning is proven to converge under reasonable assumptions \cite{Tsitsiklis1994}.
 \emph{Deep Q-Learning (DQN)} \cite{DBLP:journals/corr/MnihKSGAWR13} is a popular approach to generalize Q-Learning to high-dimensional environments. DQN approximates the Q-function by a neural network parameterized by $\theta$. At every time step $t$, the $(s_t, a_t, r_t, s_{t+1})$ experience tuple is added to an experience buffer $\mathcal{D}$ and the Q-network is optimized on the loss $L_t(\theta_t)$ computed on a mini-batch of experiences: %
$$L_t(\theta_t) = E_{(s_i,a_i,r_i,s_{i+1}) \sim \mathcal{U(D)}} \big[ (y_i(s_i, a_i) - Q(s_i, a_i ; \theta_t))^2 \big] $$
with $y_i(s_i, a_i) = r_i + \gamma \max_{a'} Q(s_{i+1}, a' ; \theta^-_t)$ and $\theta^-$ the parameters of the \emph{target network}. Training towards a fixed target network prevents approximation errors from propagating too quickly from state to state, and sampling experiences to train on (\emph{experience replay}) increases sample efficiency and reduces correlation between training samples.
\emph{Prioritized experience replay} \cite{dblp:journals/corr/schaulqas15} improves training time by sampling transitions with large residual errors from which the agent can learn more.
\subsection{Multi-objective RL}
Multi-Objective MDPs (MOMDP) \cite{WhiteKim80} are MDPs with a vector-valued reward function ${\bf r}_t = {\bf R}(s_t, a_t)$. %
Each component of ${\bf r}_t$ corresponds to one objective.
A scalarization function $f$ maps the multi-objective value ${\bf V}^\pi$ of a policy $\pi$ to a scalar value, i.e., the user utility. In this paper we focus on linear $f$; each objective, $i$, is given a weight $w_i$, such that the scalarization function becomes $f({\bf V}^{\pi},{\bf w}) = {\bf w} \cdot {\bf V}^{\pi}$.
An optimal solution for an MOMDP under linear $f$ is a \emph{convex coverage set (CCS)}, i.e., a set of undominated policies containing at least one optimal policy for any %
linear scalarization \cite{roijers:2013:sms:2591248.2591251}. Depending on whether the focus is on asymptotic \cite{jmlr09-taylor} or cumulative performance, we distinguish Offline and Online Multi-Objective RL (MORL). In this paper we focus on Online MORL.

\textbf{Offline MORL} \;\;
To learn vector-valued Q-functions for a given $\bf w$, \emph{scalarized deep Q-learning (SDQL)} \cite{DBLP:journals/corr/MossalamARW16} extends the DQN algorithm to MORL, by modifying the loss: %
$$L_t(\theta_t) = E_{(s_i,a_i,r_i,s_{i+1}) \sim \mathcal{U(D)}} \big[\frac{1}{N}\mathds{1} \cdot ({\bf y}_i - {\bf Q}(s_i, a_i ; \theta_t))^2 \big] $$
with 
${\bf y}_i\!\!=\!\!{\bf r}_i + \gamma {\bf Q}(s_{i+1}, \argmax_{a'} [{\bf Q}(s_{i+1}, a' ; \theta_t^-)\cdot {\bf w}],\theta_t^-)$.

By sequentially training Q-networks until convergence on corner weights, they approximate the CCS with a set of Q-networks.

\textbf{Online MORL} \;\;
Offline methods can be undesirable. In the \emph{dynamic weights setting} for example, the weights of the scalarization function $f$ can vary over time, and there is often not enough time to learn an entire CCS beforehand. Furthermore, the performance is evaluated with regards to the cumulative regret, i.e., the cumulative difference between the value the optimal policy would have obtained and the actual performance of the agent. In this setting, pre-training is not adequate, as it requires spending a lot of time training in anticipation rather than on the active weight vectors. %
Instead, the agent should learn, remember and apply policies on-the-fly as the weight vector changes.
In tabular RL, \citeauthor{Natarajan:2005:DPM:1102351.1102427} (\citeyear{Natarajan:2005:DPM:1102351.1102427}) have shown that instead of restarting training from scratch every time $\bf w$ changes, it is highly beneficial to continue learning from a previously learned policy. When the weight vector $\bf w$ changes to another value $\bf w'$, the policy $\pi$ that was learned for $\bf w$ is stored in a set of policies $\Pi$, along with its value vector ${\bf V}^{\pi}$. As an initial policy for the new weight vector $\bf w'$, they select from $\Pi$ the past policy with the highest scalarized value; $\pi_{init} = \argmax_{\pi \in \Pi} {\bf V}^\pi \cdot {\bf w'} $. 

\textbf{Universal Value Function Approximators (UVFA)} \;\;
\citeauthor{pmlr-v37-schaul15} \citeyear{pmlr-v37-schaul15} build a single network capable of generalizing over multiple goals. Based on the observation that a goal is often a subset of the set of states, the network learns goal and state embeddings and uses a distance-based metric to combine both embeddings. This is achieved offline, by learning several value functions independently, factorizing embeddings and then training a network to approximate these values for any given goal. 
In MORL, a goal would be a specific weight vector and as such there is no clear relation between the goal (i.e., the importance of each objective) and the state. One could fall back on the concatenation of state and goal embeddings as suggested in \cite{pmlr-v37-schaul15}, but they found this %
is prone to instability.

\section{Contributions}

Existing Deep (MO)RL algorithms are insufficient in the dynamic weights setting because they either build a complete set of policies in advance or spend a long time adapting to weight changes.
 We first propose our \emph{Conditioned Network} method capable of generalizing multi-objective Q-values across weight vectors and then we propose \emph{Diverse Experience Replay} to improve sample efficiency and counter the replay buffer's bias to recent weight vectors. %
\begin{figure}
	\center
	\includegraphics[width=.4\textwidth]{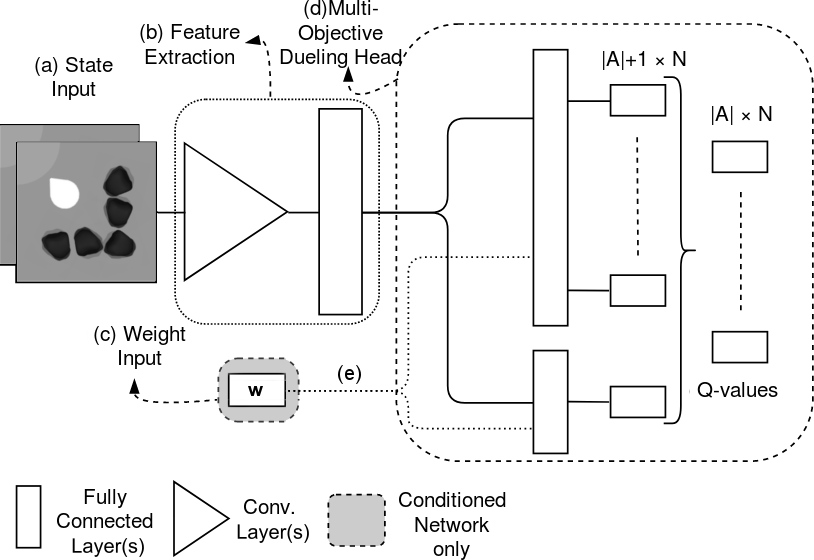}
	\caption{Features are extracted from the raw input by convolutional layers followed by a fully connected layer. The extracted features (output of (b)) are fed into an N objectives Dueling DQN head (d). The conditioned architecture feeds a weight input (c) into the Q-value head (link (e)).}
	\label{fig:architecture}
\end{figure}

\subsection{Conditioned Network (CN)}

We propose our first main contribution, Conditioned Network (CN), 
in which a UVFA is adapted to output Q-value-vectors conditioned on an input weight vector (Figure \ref{fig:architecture}). The training algorithm follows the standard DQN algorithm, i.e., the agent acts $\varepsilon$-greedily and stores its experiences in a replay buffer from which transitions are sampled to train the network on. Because the network also takes a weight-vector as input, the selection of weight vectors to train the network on requires additional consideration.

While generalization is important, attention should be paid to the active weight vector, such that the agent can quickly perform well for the objectives that are important at the moment. 
	 However, if we do not maintain trained policies, the network may overfit to the current region of the weight space and forget past policies.
To avoid this overfitting, we propose that samples should be trained on more than one weight vector at a time. 
Specifically, to promote quick convergence on the new weight vector's policy and to maintain previously learned policies, 
each experience tuple in a mini-batch is updated w.r.t.\ the current weight vector and a random previously encountered weight vector. Given a mini-batch of $B$ transitions, we compute the loss for a given transition  $(s_j,a_j,{\bf r}_j,s_{j+1})$ as the sum of the loss on the active weight vector ${\bf w}_t$ and on ${\bf w}_j$ randomly sampled from the set of encountered weights.
$$\frac{1}{2}\big[|{\bf y}^{(j)}_{{\bf w}_t}-{\bf Q}_{CN}(a_j,s_{j};{\bf w}_t)|+|{\bf y}^{(j)}_{{\bf w}_j}-{\bf Q}_{CN}(a_j,s_{j};{\bf w}_j)|\big]$$
$${\bf y}^{(j)}_{\bf w} = {\bf r}_j + \gamma {\bf Q}_{CN}^-(\argmax_{a\in A} {\bf Q}_{CN}(a,s_{j+1};{\bf w}) \cdot {\bf w},s_{j+1};{\bf w})$$
where ${\bf Q}_{CN}(a,s;{\bf w})$ is the network's Q-value-vector for action $a$ in state $s$ and weight vector ${\bf w}$. 
Training the same sample on two different weight vectors has the added advantage of forcing the network to identify that different weight vectors can have different Q-values for the same state.
Please see Appendix 1.3 for a detailed description of the CN algorithm.

This method deviates from UVFA on three major points. First, the outputs of our network are multi-objective, secondly, the whole network is trained end-to-end, and finally, we stabilize learning through our update rule which is adapted to the dynamic weights setting.
\subsection{Diverse Experience Replay}
A particular challenge to using experience replay for Deep MORL is that an experience buffer obtained through a weight vector's optimal policy can be harmful to another weight vector's training process. Existing offline %
approaches circumvent this %
by resetting the replay buffer when the trained policy changes and restarting the exploration phase.
However, excessive exploration harms cumulative performance in the (online) dynamic weights setting.
To ensure the agent learns adequately, the replay buffer must contain experiences relevant\footnote{Relevant experiences are experiences that can be expected to help a learner converge towards the optimal policy.} to any future weight vector's optimal policy. This not the case when using a standard replay buffer, as it is biased towards recently encountered weight vectors. A policy $\pi^{\bf w}$ trained exclusively on experiences obtained through another policy $\pi^{\bf w'}$ will typically diverge from the optimal policy for $\bf w$.
Making the replay buffer larger such that early experiences obtained through random exploration are still present is impractical for two reasons; (1) unless the replay buffer is infinite, older experiences could still be erased before reaching areas of the weight-space which need them. And (2), even if these relevant experiences are still present, they could be vastly outnumbered. %
Therefore, we propose a different solution to consistently provide relevant experiences to a learner for any weight vector.

We propose \emph{Diverse Experience Replay (DER)}, a diverse buffer from which relevant experiences can be sampled for weight vectors whose policies have not been executed recently.
DER replaces %
standard recency-based replay by diversity-based memorization.
Furthermore, instead of considering each transition independently, DER handles trajectories as atomic units.
To understand why, consider a trajectory of experiences from initial state to terminal state. The absence of an experience between initial and terminal state can make the task of propagating Q-values from the terminal to the initial state infeasible, as the learner has to infer the missing link.
When using standard replay buffers this is not an issue, as experiences are added and removed sequentially. Hence, for the vast majority of experiences in standard replay buffers, the preceding and subsequent transitions are also present.
To avoid %
partial trajectories, we treat trajectories as atomic units when considering them for addition in, or deletion from the diverse buffer.
To reduce the computational cost of comparing trajectories, we compute a signature for each trajectory on which we enforce diversity. This signature can for example be the trajectory's discounted cumulative reward (i.e., its return), or the set of perceptual hashes \cite{zauner2010implementation} of the trajectory's frames.
Specifically, when a new trajectory is considered for addition into a diverse buffer $\mathcal{D}'$, each trajectory's signature is computed by a signature function ${\bf s}$. A diversity function $d$ then computes the relative diversity of each trajectory's signature. The new trajectory is only added to the diverse buffer if its inclusion %
increases the overall diversity of the diverse buffer. When it is full, the traces that contribute least to diversity are ejected from the diverse buffer.

\textbf{Diversity in the Dynamic Weights Setting} \;\;
In this setting we wish to maintain a set of trajectories relevant to any region of the weight-space by ensuring trajectories with a wide variety of future rewards are present. To achieve this, we propose to; (1) treat an episode's transitions as one trajectory, (2) use a trajectory's return vector as its signature ${\bf s}(\tau) = \sum_{t=0}^{|\tau|} \gamma^t {\bf r}_t$ and (3) use a metric from multi-objective evolutionary algorithms, called the \emph{crowding distance} \cite{nsgaii}, as a diversity function.
Applied to return vectors, the crowding distance promotes the presence of trajectories spread across the space of returns.

We also maintain a standard FIFO buffer to which experiences are added first. When this buffer is full, the oldest trajectory in it is removed and considered for addition in the diverse buffer. These two %
buffer types allow a new weight vector's policy to be bootstrapped on experiences from the diverse buffer and then further trained on the experiences it progressively adds to the standard buffer.
Please see Appendix 1.5 for a detailed description of the DER algorithm.

\subsection{Minecart Problem}

\begin{figure}[t!]
	\center
	\includegraphics[width=.14\textwidth]{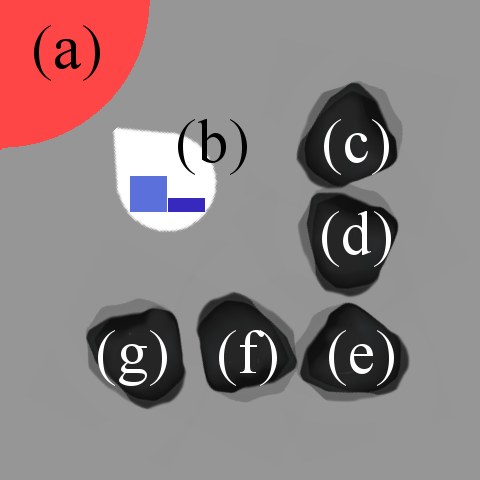}~~~
	\includegraphics[width=.16\textwidth]{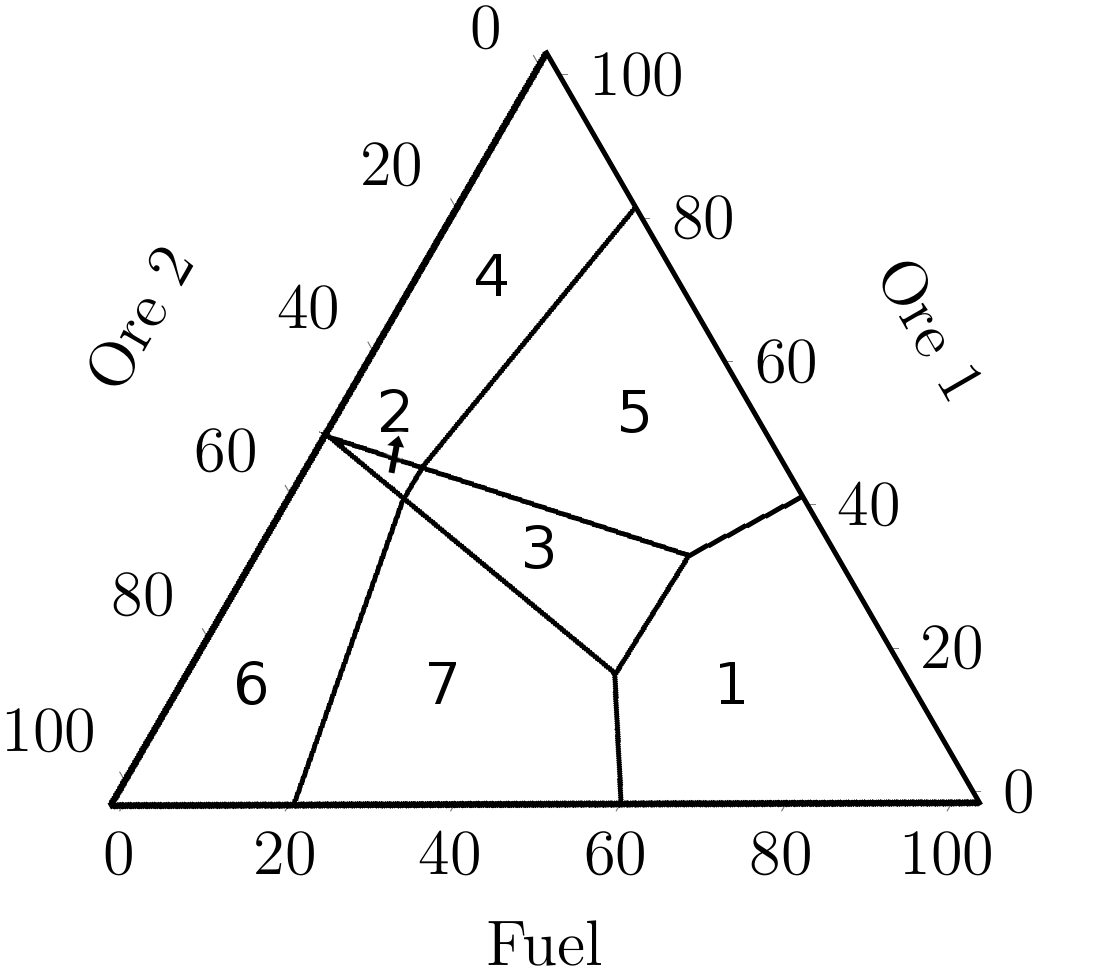}
	\caption{\textbf{Left:} Instance of the Minecart environment with 5 mines ((c) to (g)) containing varying amounts of 2 ores. The 2 bars on the minecart (b) indicate how much of each ore is present in the cart. Ores are sold on the base (a). \textbf{Right:} Weight vectors in the same region share the same optimal policy. Axes are the relative importance in \% of each objective. We distinguish (1) collecting no resources if the fuel cost is too high, (6,7) privileging ore 2, (4,5) privileging ore 1, and (2,3) privileging the quick collection of either ore. Differences between each pair lie in a higher fuel cost, in which case it is optimal to accelerate less.}
	\label{fig:minecart}
\end{figure}
Existing Deep MORL problems, such as the image version of Deep Sea Treasure (DST) \cite{DBLP:journals/corr/MossalamARW16}, are relatively trivial, i.e., it has  $4$ actions and even though the states are presented as an image, the number of actually distinct states is only around ${\sim}50$. This is in stark contrast with single-objective Deep RL for which among others, ALE \cite{DBLP:journals/corr/abs-1207-4708} provides a diverse set of challenging environments. To close this gap, we propose an original %
benchmark, the Minecart problem\footnote{The code can be found at \url{https://github.com/axelabels/DynMORL}}. %
Minecart %
has a continuous state space, stochastic transitions and delayed rewards. %
The Minecart environment consists of a rectangular image, depicting a base, mines and the minecart controlled by the agent.
A typical frame of the Minecart environment is given in Figure \ref{fig:minecart} Left. Each episode starts with the agent on top of the base. Through the \textit{accelerate}, %
\textit{brake}, %
\textit{turn left}, %
\textit{turn right}, %
\textit{mine}, or \textit{do nothing} %
actions, 
the agent should reach a mine, collect resources %
and return to the base to sell them. %

The reward vectors are N-dimensional: ${\bf r}=(r_1,...,r_N)$. The first $N\!-\!1$ elements correspond to the amount of each of the $N\!-\!1$ resources the agent sold, the last element is the consumed fuel. Particular challenges of this environment are the sparsity of the first $N\!-\!1$ components of the reward vector, %
as well as the delay between actions (e.g., mining) and resulting rewards. %
The resources an agent collects by mining are generated from the mine's random distribution, resulting in a stochastic transition function. All other actions are deterministic.
The weight vector ${\bf w}$ expresses the relative importance of the objectives, i.e., the price per resource. %
For the default configuration of the Minecart, the weight-space has 7 regions with a different optimal policy (Figure \ref{fig:minecart}, right). A full description of the environment %
and default parameter values is given in the appendix.
In the dynamic weights Minecart problem,  an agent should quickly adapt to fluctuations in the price of resources.%

\section{Adapted Algorithms}

For %
completeness, we show how methods from related settings can be adapted to dynamic weights, but are suboptimal.

\subsection{UVFA}
We first present how we adapted UVFA. To avoid expensive pre-training, we consider as basis the direct bootstrapping variant of UVFA and train the network end-to-end. Because a distance based metric is not applicable in our setting\footnote{In our case, the goal (a specific $\bf w$) is not comparable to a subset of states.}, we concatenate the state features and the goal (i.e., weight-vector) and feed them into the policy heads. The network thus shares the same overall architecture as CN but outputs scalar Q-values.
Following UVFA's direct bootstrapping, we sample goals and transitions from the replay buffer to train the network on. For each transition $(s_j,a_j,{\bf r}_j,s_{j+1})$ of a mini-batch, we sample ${\bf w}_j$ from the set of encountered weights and minimize the loss
$|y_j-{ Q}(a_j,s_{j};{\bf w}_j)|$, with 
$$y_j = {\bf r}_j \cdot {\bf w}_j + \gamma { Q}^-(\argmax_{a\in A} {Q}(a,s_{j+1};{\bf w}_j),s_{j+1};{\bf w}_j)$$
 ${ Q}(a,s;{\bf w})$ is the network's Q-value for action $a$ in state $s$ and weight vector ${\bf w}$. Setting $g\!=\!\bf w_j$ and replacing ${\bf r}_j \cdot {\bf w}_j$ by $R_g(s_j,a_j,s_{j+1})$ in the above equations gives an equivalent goal-oriented notation as in \cite{pmlr-v37-schaul15}.

\subsection{Multi-Network (MN)}
Combining existing work on tabular dynamic weights \cite{Natarajan:2005:DPM:1102351.1102427} and multi-objective deep RL for different settings \cite{DBLP:journals/corr/MossalamARW16}, we propose to gradually build a set of policies represented by MO Q-networks, $\Pi$. Key insights of this approach are that; (1) for a given ${\bf w}$ we can train a Q-network for a region of the weight-space around ${\bf w}$, (2) by training multiple Q-networks on different weight vectors we can cover more regions of the weight-space, and (3) we can speed up learning by knowledge transfer from previously trained neural networks.

By only storing un-dominated Q-networks (i.e., Q-networks that are optimal for at least one encountered weight vector\footnote{By encountered weight vectors we mean the set of weight vectors the agent has experienced since it started learning.}), we gradually approximate $\Pi$, a subset of the CCS relevant to the encountered weights.
Because a CCS is typically relatively small, the number of networks we need to train and maintain in memory is also expected to be small.

Each policy  $\pi_{{\bf w}}$ is trained for the active weight vector ${\bf w}$ following \emph{scalarized deep Q-learning} \cite{DBLP:journals/corr/MossalamARW16}.
 When the active weights change, the stateless value of the policy $\pi_{{\bf w}}$, ${\bf V}^{\pi_{\bf w}}$, is compared to all previously saved policies.
If ${\bf V}^{\pi_{\bf w}}$ improves upon the maximum scalarized value of the policies already in $\Pi$ for at least one past weight vector or for the current weight vector $\bf w$, it is saved, otherwise it is discarded. To limit memory usage and ensure fast retrieval by keeping $\Pi$ small, all old policies made redundant by ${\pi_{\bf w}}$ are removed from $\Pi$. A policy is redundant if it is not the best policy for any encountered weight vector. 
We hot-start learning for each new $\bf w$ by copying the policy $\pi' \in \Pi$ whose scalarized value ${\bf V}^{\pi'} \cdot {\bf w}$ is maximal.
Following previous transfer learning approaches \cite{DBLP:journals/corr/MossalamARW16,DBLP:journals/corr/ParisottoBS15,2016DeepRL-Du}, MN copies parameters from a source network ($\pi'$'s Q-network) to the current policy's Q-network. 
Because MN compares policies based on predicted Q-values, inaccurate outputs disturb training by biasing MN to overestimated policies. As a result, MN needs long training times for each weight vector to obtain accurate values to compare.
Please see Appendix 1.4 for a detailed description of the MN algorithm.

\section{Experimental Evaluation}

We test the performance of our algorithms on two different problems: the image version of Deep Sea Treasure (DST) proposed by \citeauthor{DBLP:journals/corr/MossalamARW16} (\citeyear{DBLP:journals/corr/MossalamARW16}), and our newly proposed benchmark, the Minecart problem. 
Moreover, we use two weight change scenarios. 
We first evaluate the performance when weight changes are sparse, as in \cite{Natarajan:2005:DPM:1102351.1102427}, in which case an agent (and its replay buffer) could overfit to the active weights. Then, we look at %
regular weight changes, in which case it can be tempting to learn a policy that is good for most weights but optimal for none. We compare CN against MN, UVFA, %
a Multi-Objective DQN trained on the current $\bf w$ only (MO), and two ablated versions of CN, CN-ACTIVE and CN-UVFA.

\subsection{Experimental Setup}
First, we evaluate the performance for \emph{sparse} and large weight changes; the current weight, $\bf w$, is randomly sampled from a Dirichlet distribution ($\alpha=1$) every $50k$ steps for Minecart and $5k$ steps for DST.
Second, we test on \emph{regular} weight changes; $\bf w$ linearly moves to a random target, $\bf w'$, over 10 episodes, after which a new $\bf w'$ is sampled.
Both variants are evaluated on the Minecart environment, and on an image version of Deep Sea Treasure (DST, fully described in the appendix).%

We evaluate policies based on their \emph{regret}, i.e., the difference between optimal value and actual return,
$\Delta({\bf g,w})\!=\!{\bf V^*_w}\!\cdot\!{\bf w} - {\bf g\!\cdot\!w}  =  {\bf V^*_w}\!\cdot\!{\bf w}  - \sum_{t=0}^T \gamma^t {\bf r_t}\!\cdot\!{\bf w} $,
where $\bf g$ is the discounted cumulative reward, $\bf V^*_w$ denotes the optimal value for $\bf w$, $\{{\bf r}_0,...,{\bf r}_T\}$ is the set of vector-valued rewards collected during an episode of length $T$.
Unlike the return, the regret allows for a common optimal value regardless of the weights, i.e., an optimal policy always has $0$ regret. This is a necessary condition to consistently evaluate performance over different runs and for different weight vectors.

We include the performance of the adapted algorithms we proposed, the MN algorithm as well as UVFA.
To show the benefits of our proposed loss for CN we perform an ablation study by also (1) training only on the active weight vector (CN-ACTIVE) and (2) training only on randomly sampled weight vectors (CN with UVFA loss, CN-UVFA). 
As a baseline, we use a basic \emph{Multi-Objective DQN approach (MO)}; a single multi-objective DQN continuously trained on only the current $\bf w$ through scalarized Deep Q-learning. MO does not maintain multiple networks and the weight vector is not fed as input to the network.
An alternative naive baseline for general MORL purposes suggested by \cite{6918520} learns optimal Q-values for each objective then selects actions by scalarizing these multiple single-objective Q-values. Because the resulting Q-value-vectors do not capture the necessary trade-offs, this baseline can only perform in edge cases where one objective outweighs all others. As a result it performed poorly in our tests and we restrict its experimental results to the appendix.

All algorithms are run with and without DER and with prioritized sampling 
\cite{dblp:journals/corr/schaulqas15}.

\subsection{Results}
Results for each weight change scenario are collected over 10 runs. Plots are smoothed by averaging over 200 steps.
\begin{figure*}[t]
	\centering
	\begin{subfigure}[t]{0.32\textwidth}
		\includegraphics[width=\textwidth, trim=.2mm 72mm 0mm 72mm, clip=true]{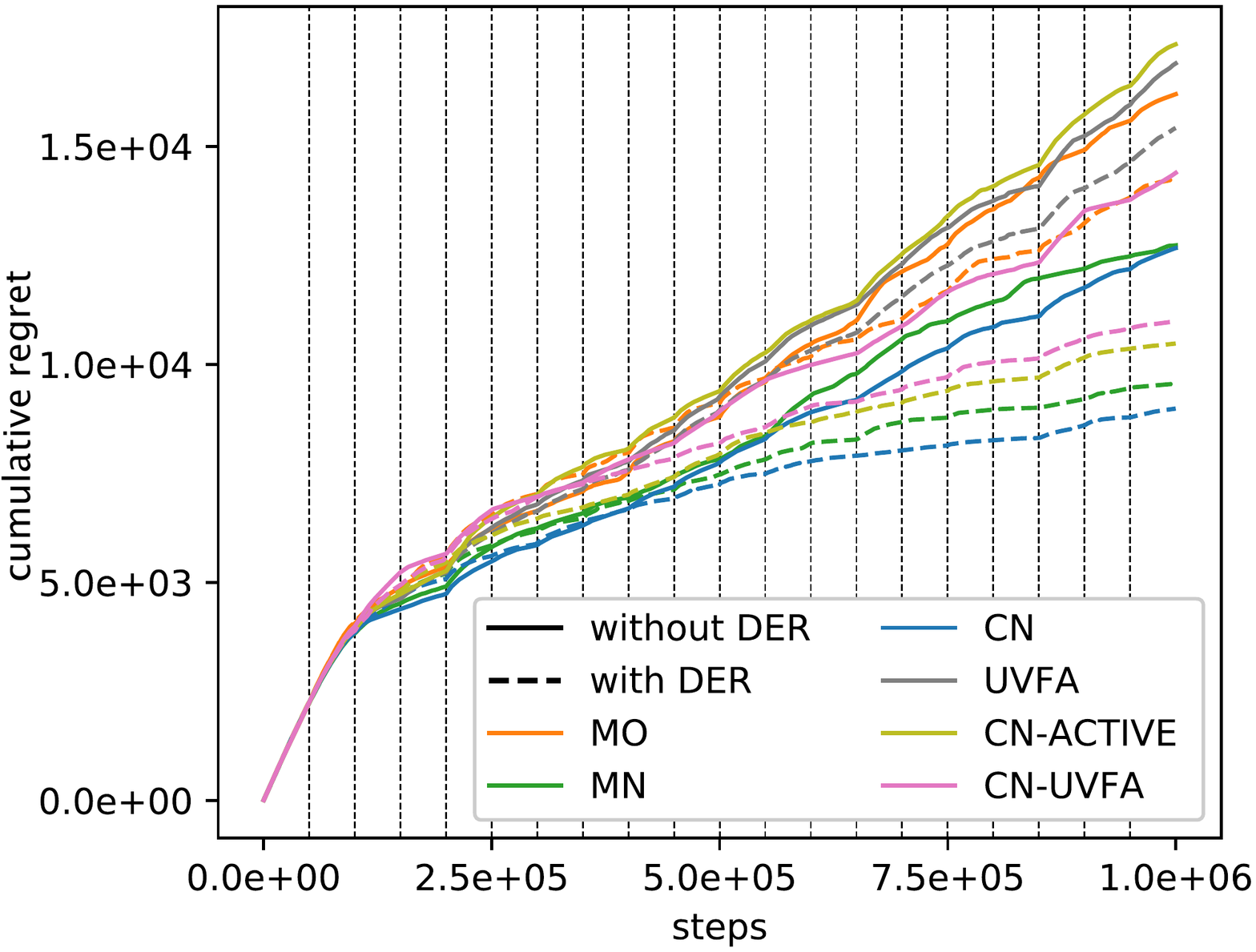}
	\end{subfigure}%
	~ 
	\begin{subfigure}[t]{0.25\textwidth}
		\centering
		\includegraphics[width=\textwidth, trim=45mm 73mm 27mm 80mm, clip=true]{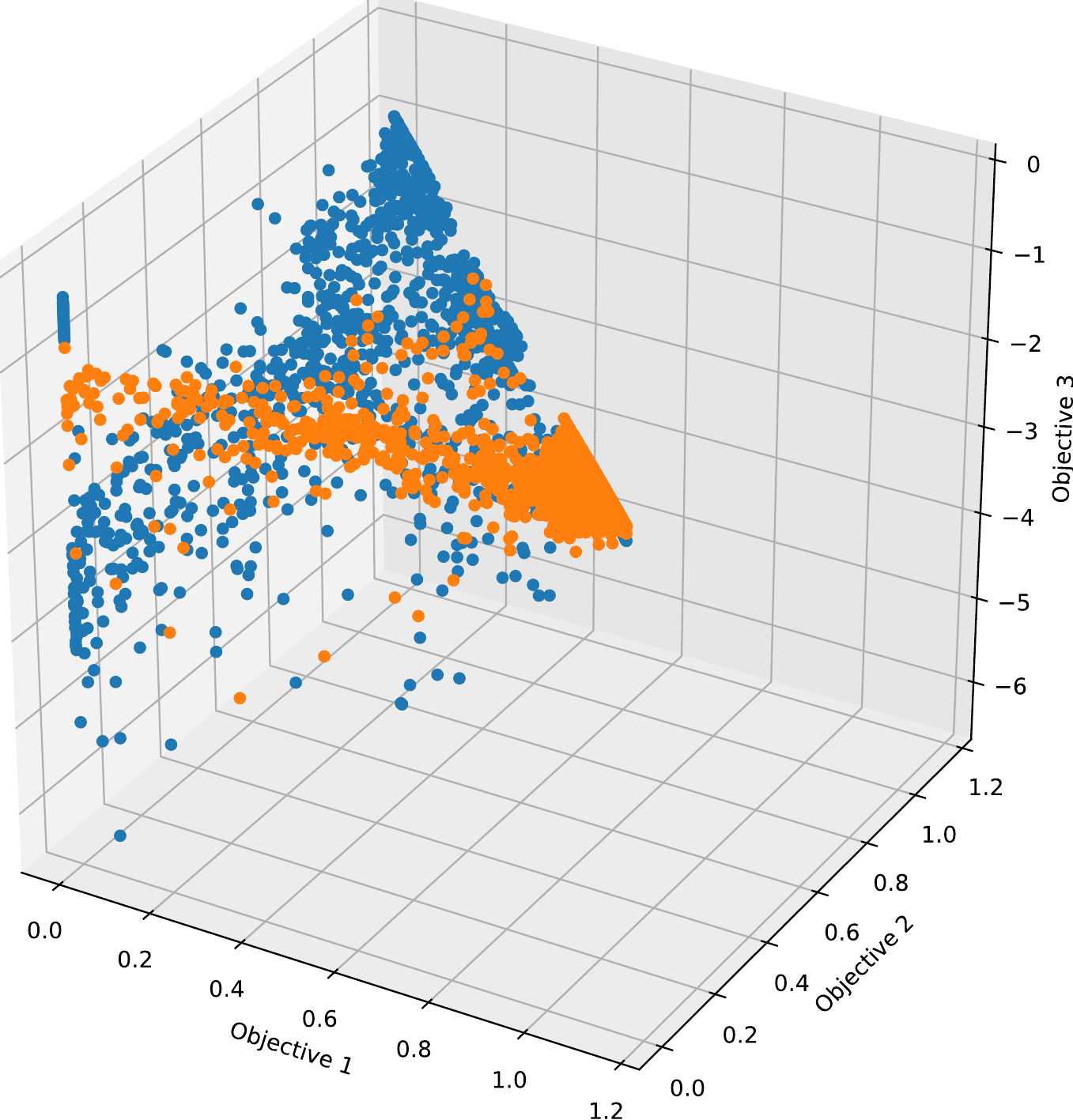} 
	\end{subfigure}
	~
	\begin{subfigure}[t]{0.32\textwidth}
		\includegraphics[width=\textwidth, trim=0mm 72mm 0mm 72mm, clip=true]{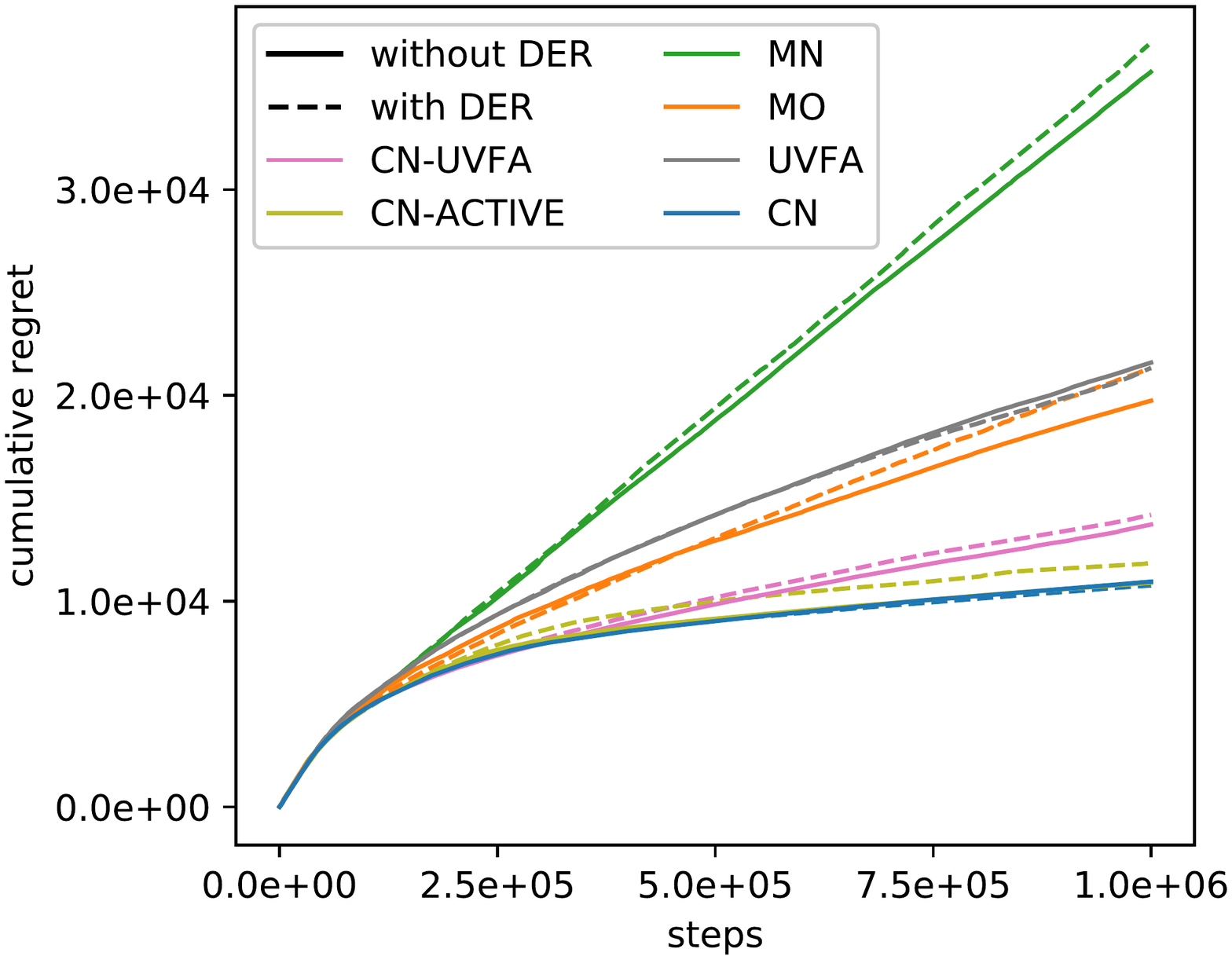}
	\end{subfigure}
	\caption{Solid lines plot performance without DER, dashed lines plot the performance with DER. \textbf{Left:} Cumulative regret for the Minecart problem when weights change every 50k steps (vertical lines), MN+DER and CN-UVFA+DER overlap eachother. \textbf{Middle:} Effect of DER on the replay buffer's content, each dot represents a trajectory's return vector. The non-diverse buffer (orange dots) is biased towards the recent weight-vector (favoring objective 1). The diverse buffer (blue dots) maintains a set of returns spread across the space of possible returns. \textbf{Right:} Cumulative regret for the Minecart problem when weights change over the span of 10 episodes, CN, CN+DER and CN-ACTIVE overlap in the lowest curve.} \label{fig:mc-plots}
\end{figure*}
\begin{table*}[t]
	\caption{Average episodic regret (Mean $\Delta$) and improvement over MO with Standard ER baseline ($>$baseline) for both weight change scenarios (lower is better). %
		We distinguish overall performance and performance over the last 250k steps.}\label{table:results}
	\small
	\centering
	\begin{tabular}{|c|c|l|l|l|l|l|l|l|l|}

		\cline{3-10}

		\multicolumn{2}{c|}{}& \multicolumn{4}{|c|}{Overall} & \multicolumn{4}{|c|}{Last 250k steps}\\

		\cline{3-10}

		\multicolumn{2}{c|}{}& \multicolumn{2}{|c|}{Standard ER}       & \multicolumn{2}{|c|}{DER}               & \multicolumn{2}{|c|}{Standard ER}       & \multicolumn{2}{|c|}{DER}        \\

		\cline{2-10}

		\multicolumn{1}{c|}{}& Algorithm & Mean $\Delta$  & $>$baseline      & Mean $\Delta$          & $>$baseline               & Mean $\Delta$ & $>$baseline       & Mean $\Delta$           & $>$baseline                \\
		
		\cline{2-10}

		\hline
		
		& MO & 0.324 & $-$ & 0.285 & -12.04\% & 0.275 & $-$ & 0.207 & -24.73\% \\

		\cline{2-10}
		Sparse & MN & 0.255 & -21.3\% & 0.191 & -41.05\% & 0.139 & -49.45\% & \textbf{0.063} & \textbf{-77.09\%} \\

		\cline{2-10}
		Weight & CN & 0.253 & -21.91\% & \textbf{0.18} & \textbf{-44.44\%} & 0.184 & -33.09\% & 0.068 & -75.27\% \\
		Changes & CN-UVFA & 0.288 & -11.11\% & 0.22 & -32.1\% & 0.218 & -20.73\% & 0.102 & -62.91\% \\
		& CN-ACTIVE & 0.347 & +7.1\% & 0.21 & -35.19\% & 0.316 & +14.91\% & 0.088 & -68.0\% \\

		\cline{2-10}
		& UVFA & 0.338 & +4.32\% & 0.308 & -4.94\% & 0.302 & +9.82\% & 0.253 & -8.0\% \\

				\hline
				
		& MO & 0.398 & $-$ & 0.43 & +8.04\% & 0.258 & $-$ & 0.319 & +23.64\% \\
		\cline{2-10}
		Regular & MN & 0.718 & +80.4\% & 0.746 & +87.44\% & 0.67 & +159.69\% & 0.709 & +174.81\% \\
		\cline{2-10}
		Weight & CN & 0.222 & -44.22\% & \textbf{0.219} & \textbf{-44.97\%} & 0.069 & -73.26\% & \textbf{0.064} & \textbf{-75.19\%} \\
		Changes & CN-UVFA & 0.278 & -30.15\% & 0.287 & -27.89\% & 0.149 & -42.25\% & 0.149 & -42.25\% \\
		& CN-ACTIVE & 0.221 & -44.47\% & 0.24 & -39.7\% & 0.065 & -74.81\% & 0.071 & -72.48\% \\
		\cline{2-10}
		& UVFA & 0.435 & +9.3\% & 0.43 & +8.04\% & 0.273 & +5.81\% & 0.267 & +3.49\% \\

				\hline

	\end{tabular}
	
\end{table*}
\begin{figure}[h!]
	\centering
	\begin{subfigure}[t]{0.215\textwidth}
		\includegraphics[width=\textwidth, trim=0.2mm 72mm 0mm 72mm, clip=true]{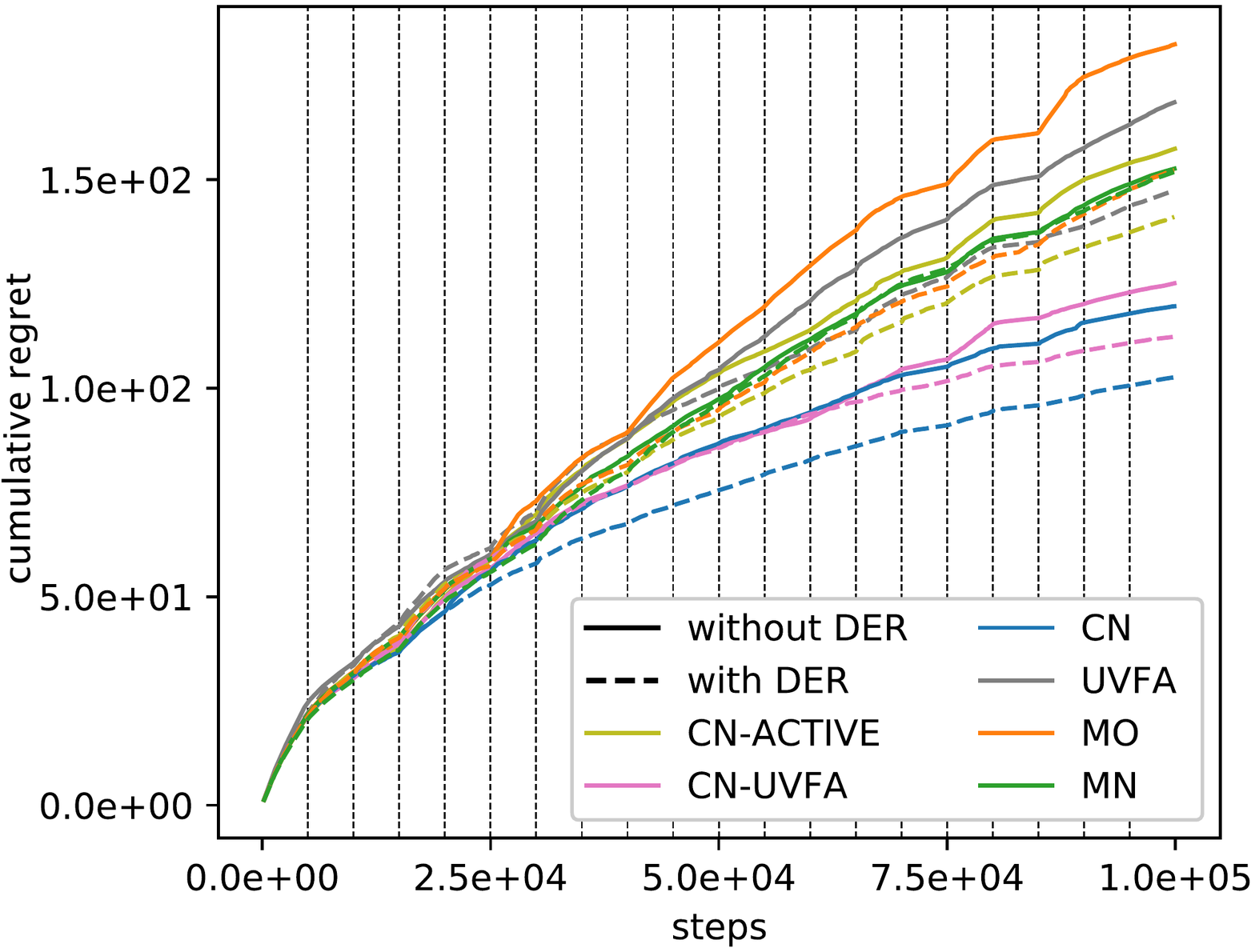}
		
	\end{subfigure}%
	~
	\begin{subfigure}[t]{0.215\textwidth}
		\includegraphics[width=\textwidth, trim=0mm 72mm 0mm 72mm, clip=true]{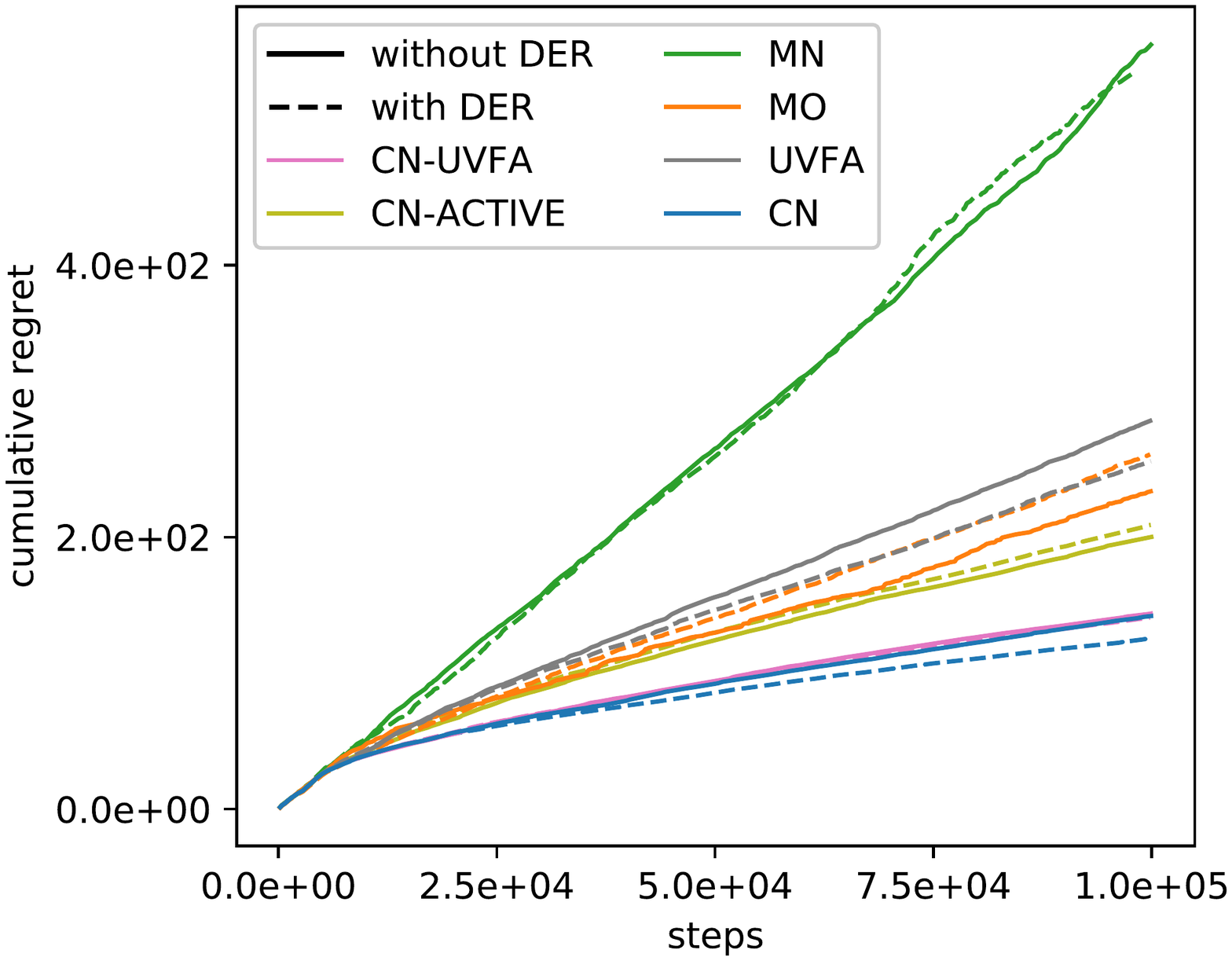}
		 
	\end{subfigure}
	\caption{Cumulative regret for DST. Solid lines represent performance without DER, dashed lines with DER. \textbf{Left:} sparse  weights changes, every 5K steps (vertical lines). \textbf{Right:} regular weight changes over the span of 10 episodes. CN and CN-UVFA(+DER) overlap near the bottom.}\label{fig:dst-plots}
\end{figure}
\subsubsection{Sparse weight changes}

We determine how robust our algorithms are to overfitting to recent weight vectors by evaluating the performance for few but large weight changes  (Left plots, Figures \ref{fig:mc-plots} and \ref{fig:dst-plots}). Here, the main challenges %
are that (1) the agent's policy could overfit to the current $\bf w$ and forget policies for past weight vectors, and (2) the replay buffer could be biased towards experiences for recent $\bf w$'s.

\textbf{Minecart} \;\;
As the MO baseline is unable to remember previously learned policies, it must repeatedly (re-)learn policies, leading to a loss in performance whenever weight changes occur. Moreover, %
the replay buffer bias prevents the MO agent from efficiently converging to new optimal policies. %
Using DER %
helps in this respect.
The middle plot in Figure \ref{fig:mc-plots} illustrates the effect of having a secondary diverse buffer (DER). While recent experiences (orange) are concentrated in the same region, the diverse experiences (blue) are spread across the space of possible returns.
By storing trained policies, MN can continue learning from the best policy in memory for each new $\bf w$. %
However, if no relevant experiences are in the buffer, it can be unable to optimize for the new weights. Thus, as for MO, the inclusion of DER significantly improves performance. %
While MN learns more slowly than other algorithms %
it is on par with the best performing algorithms over the last 250k steps when using DER, as it takes time to train a suitable set of policies.
Comparing CN to MN, we find that their performance is similar without DER.
In addition to the difficulty of learning a new policy without diversity, CN is also susceptible to forgetting learned policies if the replay buffer is biased towards another policy. DER solves both problems and significantly improve performance.
We find that over the last 250k steps, CN's performance with DER is not significantly different from MN's performance with DER. However, overall performance does improve over MN. We thus conclude that while MN and CN both ultimately learn a good set of policies, CN does so quicker.
Additionally, we find that CN with our proposed loss outperforms the alternatives. Specifically, training uniformly on weight vectors sampled from the set of encountered weight vectors (CN-UVFA) significantly hurts performances without DER. By not consistently training on the current $\bf w$, CN-UVFA puts more effort into maintaining old policies than into learning new policies.  %
As a result, it takes longer for the agent to perform well for new weight vectors, %
and relevant experiences are less likely to be collected. Hence, slower convergence leads to fewer relevant experiences, in turn leading to slower convergence. When we include DER, relevant experiences are present despite the slower convergence, leading to a smaller impact on performance. UVFA shares the same flawed weight selection as CN-UVFA, and in addition only outputs scalar Q-values, meaning it does not exploit the added structure provided by the multi-objective rewards. These two factors in combination with the single-goal loss lead to performance close to our MO baseline.
When we only train the Conditioned Network on the active weight vector (CN-ACTIVE), there is no explicit mechanism to preserve past policies, as a result CN-ACTIVE is likely to overfit to the current $\bf w$. CN-ACTIVE is outperformed in overall and final performance by MN, CN as well as by MO and CN-UVFA without DER. 

\textbf{DST} \;\;
We find that, while CN+DER still performs best for the DST problem, the performance of other algorithms is permuted. While the relative performances of MN, CN and CN-UVFA seem similar to those we obtained for the minecart problem, we found that CN-ACTIVE and MN perform relatively worse. What's more, DER seems to have no significant impact on the performance of MN. We hypothesize that MN performs worse for DST than for Minecart because the smaller distance between optimal policies in DST is harder to distinguish from approximation errors. 

\subsubsection{Regular weight changes}
When weights change quickly, agents could fail to converge in time, resulting in %
sub-optimal policies for most weights.

\textbf{Minecart} \;\;
As the rightmost plot in Figure \ref{fig:mc-plots} illustrates, regular weight changes lead to significantly different results w.r.t. sparse weight changes.
While there is a slight loss in performance when adding DER to MO, there is no qualitative difference, as we found that with or without DER, MO converges to a single policy and applies it for all weight vectors.
In contrast, CN learns to perform close to optimally for all weight vectors. Because CN continuously trains a single network towards multiple policies, its training process is not affected by the regular changes.
In Minecart, when weights change regularly, CN-ACTIVE does not have enough time to overfit, resulting in performance on par with CN. CN-UVFA's performance remains poor, suggesting that emphasizing training on the current weight vector is crucial in Minecart. UVFA's performance is again close to MO, confirming it is not suited to the online dynamic weights setting. %
Due to the short per-weight training times, the networks in MN do not have enough time to converge for any given weight vector. As a result, their outputs, on which selection is done, are inaccurate. This makes MN discard more accurate newer policies in favor of older overestimated policies, and ultimately prevents it from learning.  %
In contrast to sparse weight changes, there is no significant benefit to using DER %
as, due to the regular small weight changes, relevant experience is still in the replay buffer for new $\bf w$. 

\textbf{DST} \;\;
We obtained similar results for CN in DST. However, CN-ACTIVE performs worse for DST (-15\%) than for Minecart (-44\%). We hypothesize that when the distance between optimal policies is large (as in Minecart) focus should be put on the active weight vector to close the gap to the new optimal policy. Conversely, when optimal policies are close together (as in DST), unmaintained policies can more easily diverge from an optimal policy to a near-optimal policy.

In summary, %
our new algorithm CN dominates all other algorithms (with and without DER). We conclude that our proposed loss %
balances between learning new policies and maintaining learned policies well. Furthermore, MN is only able to perform well when given enough training time to learn accurate Q-values. %
Finally, DER improves performance when diversity cannot be expected to occur naturally.

\section{Related Work}

\citeauthor{Natarajan:2005:DPM:1102351.1102427} (\citeyear{Natarajan:2005:DPM:1102351.1102427}) introduce the dynamic weights setting and show how it can be solved for low-dimensional problems by training a set of policies through tabular RL. The MN algorithm shares the same ideas, but addresses the additional challenges of Deep RL. Similar to MN, DOL \cite{DBLP:journals/corr/MossalamARW16} solves an image version of DST for the (off-line MORL) unknown weights scenario rather than the (on-line MORL) dynamic weights scenario. DOL builds a CCS in which each policy is implemented by a DQN. However, (1) the solved problem has a small underlying state-space while our Minecart problem is continuous, (2) weights chosen by DOL can be trained upon as long as necessary, and (3) only the final performance matters.
Selective Replay \cite{DBLP:journals/corr/abs-1802-10269} prevents catastrophic forgetting in single-objective multi-task problems. Similarly, \cite{de2018experience} propose alternatives to FIFO buffers for single-objective Deep RL. Neither solution factors in the challenges we addressed with DER (e.g., long-term dependencies between experiences which we handle by storing trajectories) and hurt performance in this setting (please see the appendix for an experimental comparison). Successor Features (SF) \cite{DBLP:journals/corr/BarretoMSS16} decomposes a scalar reward into a product of state features and task weights to enable transfer learning between tasks. While these two components are analogous to the multi-objective reward and weight vectors, our work focuses on learning when this decomposition is given rather than learning the decomposition. Removing this decomposition learning from the proposed \textit{SFQL} reduces it to an algorithm similar to \textit{MN}. Universal Successor Features Approximators \cite{borsa2018universal} and Universal Successor Representations \cite{DBLP:journals/corr/abs-1804-03758} combine the benefits of SF and UVFA to further generalize across goals. As for SF and UVFA separately, the challenges of Online MORL are not addressed.
\section{Conclusion and Future Work}
In this paper, we proposed the CN algorithm capable of tackling high-dimensional dynamic weights problems by learning weight-dependent multi-objective Q-values. %
We identified the drawbacks of FIFO experience replay for dynamic weights and proposed DER, which maintains a set of trajectories such that any policy can benefit from experiences present in this secondary buffer.
To evaluate the performance of our algorithms we introduced the high-dimensional, continuous and stochastic Minecart problem.
Our results show that CN dominates adapted algorithms from related settings in %
different weight change scenarios. Furthermore, %
our proposed loss, on the active weight vector and a random past weight vector, enables the network to generalize across weight vectors. 
On Minecart and DST %
we showed that CN always comes close to optimality, while MN fails to converge when weights regularly change.

In future work, we aim to integrate additional transfer learning techniques to further promote knowledge re-use between weight vectors. 
 Finally, we aim to explore DER variants.

\nocite{van2016deep}
\nocite{dblp:journals/corr/wangfl15}
\nocite{Vamplew2011}

\newpage

\bibliography{dyn}
\bibliographystyle{icml2019}

\twocolumn[
	\icmltitle{Supplemental Material to Dynamic Weights in Multi-Objective Deep Reinforcement Learning}
\setcounter{section}{0}
\icmlsetsymbol{equal}{*}
\icmlkeywords{Multi-Objective Reinforcement Learning, Deep Reinforcement Learning}

\vskip 0.3in
]

\section{Algorithms}

In this section of the appendix, we first present some of the recent advances in Deep RL we extended to multi-objective Deep RL and then include the pseudo-code for Conditioned Network (CN), Multi-Network (MN) and Diverse Experience Replay (DER) algorithms.

\subsection{Prioritized Sampling}
For both replay buffer types, we used proportional prioritized sampling (also referred to as Prioritized Experience Replay \cite{dblp:journals/corr/schaulqas15}). This technique replaces the uniform sampling of experiences for use in training by prioritized sampling, favouring experiences with a high TD-error (i.e., the error between their actual Q-value and their target Q-value).
To update each sample's priority in the dynamic weights setting, we observe that TD-errors will typically be weight-dependent. It follows that a priority can be overestimated if the TD-error is large for the weight on which the sample was trained but otherwise low, or it can be underestimated if the TD-error is low for the trained weight but otherwise high. In the first case, the sample is likely to be resampled quickly and its TD-error will be re-evaluated. Hence we consider the overestimation to be reasonably harmless\footnote{We further note that a given sample can only be overestimated often if it repeatedly has a large TD-error for the weights it is being trained on, in which case it should not be considered as overestimated.}. In contrast, underestimating a sample's TD-error can have a more significant impact because it is unlikely to be resampled (and thus re-evaluated) soon. To alleviate this problem, we used Prioritized experience replay's $\varepsilon$ parameter which offsets each error by a positive constant $\varepsilon$;
$p(\delta) = (\delta + \varepsilon)^\alpha$.
This increases the frequency at which low-error experiences are sampled allowing for possibly underestimated experiences to be re-evaluated reasonably often. As a result, on average, experiences that get sampled less often are samples that consistently have low TD-errors for all weight vectors used in training.

The question then remains which TD-error should be used for a given sample. For both the MO baseline and MN we update the priority w.r.t.\ the TD-error on the active weight vector.

We find this to be insufficient for CN as it trains both on the active weight vector and on randomly sampled past weight vectors. Ideally we would compute a TD-error relative not only to the active weight-vector but also all the past weight vectors. However, it would be too computationally expensive to perform a forward pass of each training sample on all encountered weights, so we only consider the two weight vectors (i.e., ${\bf w}_t$ and ${\bf w}_j$) it was last trained on.
Only using the active weight vector's TD-error to determine the priority would prevent past policies from being maintained, as their TD-error would have no influence on how often experiences are trained on. Conversely, only taking the randomly sampled weight vector in consideration could hurt convergence on the active weight vector's policy. Hence we balance current and past policies by computing the average of both TD-errors and use that value to determine the experience's priority.

\subsection{Double DQN}
Double DQN \cite{van2016deep} reduces Q-value overestimation by using the online network for action selection in the training phase. I.e.,
$y_j =r_j + \gamma Q^-(argmax_{a'}Q(a',s_{j+1}),s_{j+1})$ instead of
$y_j =r_j +\gamma max_{a'} Q^-(a',s_{j+1})$. As a result, an action needs to be overestimated by both the target and the online network to cause the feedback loop that would occur in standard DQN.
The same technique can be used in multi-objective DQNs. It is especially useful for the Multi-Network algorithm, as overestimated Q-values can have a significant impact on policy selection.

\subsection{Conditioned Network Algorithm}
The Conditioned Network (CN) algorithm (Algorithm \ref{alg:single-network}) for multi-objective deep reinforcement learning under dynamic weights, handles changes in weights (i.e., the relative importance of each objective) by conditioning a single network on the current weight vector, $\bf w$. As such, the $\bf Q$-values outputted by the network depend on which $\bf w$ is inputted, alongside the state. For an architectural overview of the networks we employed, please refer to Appendix \ref{architecture}.
\begin{algorithm}[ht!]
	\small
	\caption{Dynamic Weight-Conditioned Reinforcement Learning}\label{alg:single-network}

	\begin{algorithmic}[1]
		\STATE Define $a_{{\bf w},s}^*$ as shorthand for $argmax_{a\in A} {\bf Q}_{CN}(a,s;{\bf w}) \cdot {\bf w}$

		\STATE initialize (diverse) replay buffer $\mathcal{D}$ and unique weight history $\mathcal{W}$

		\STATE ${\bf Q}_{CN},{\bf Q}^-_{CN} \gets initializeConditionedModel()$
		\FOR {steps $t \in \{0 ... T\}$}

		\STATE ${\bf w}_t \gets getWeightVector(t)$
		\STATE add  ${\bf w}_t$ to $\mathcal{W}$

		\STATE With probability $\varepsilon$ select a random action $a_t$

		\STATE Otherwise $a_t=a^*_{{\bf w}_t,s_t}$
		\STATE Execute action $a_t$ and observe ${\bf r}_t$ and ${\bf s}_{t+1}$
		\STATE Store transition $(s_t,a_t,{\bf r}_t,s_{t+1})$ in $\mathcal{D}$
		\STATE Sample minibatch of transitions from $\mathcal{D}$
		\FOR {each sampled transition $(s_j,a_j,{\bf r}_j,s_{j+1})$}
		\STATE	${\bf w}_j$ randomly sampled from $ \mathcal{U({W})}$

		\IF {transition is terminal}
		\STATE $ y_j = y'_j = {\bf r}_j$
		\ELSE
		\STATE $y_j = {\bf r}_j + \gamma {\bf Q}_{CN}^-(a^*_{{\bf w}_t,s_{j+1}},s_{j+1};{\bf w}_t)$
		\STATE $y'_j = {\bf r}_j + \gamma {\bf Q}_{CN}^-(a^*_{{\bf w}_j,s_{j+1}},s_{j+1};{\bf w}_j)$
		\ENDIF

		\ENDFOR
		\STATE perform gradient descent step on
		$$\frac{1}{2}\big[|y_j-{\bf Q}_{CN}(a_j,s_{j};{\bf w}_t)|+|y'_j-{\bf Q}_{CN}(a_j,s_{j};{\bf w}_j)|\big]$$
		\STATE Every $N^-$ steps; ${\bf Q}_{CN}^- = {\bf Q}_{CN}$  \COMMENT {Synchronize target network}
		\STATE anneal($\varepsilon$)
		\ENDFOR
	\end{algorithmic}
\end{algorithm}

After initializing the network, the agent starts interacting with the environment. At every timestep, first a weight vector ${\bf w}_t$ is perceived, and added to the set of observed weights $\mathcal{W}$ if it is different from ${\bf w}_{t-1}$. $\mathcal{W}$ is used to sample historical weights from, so that the network keeps training with regards to both current and previously observed weights. This is necessary in order to make the network generalize over the relevant part of weight simplex. Specifically, for each gradient descent step, the target consists of two equally weighted components; one for the current weight  ${\bf w}_t$ and one randomly sampled weight from $\mathcal{W}$, ${\bf w}_j$.

While not explicitly visible in the algorithm, CN makes use of prioritized experience replay. Please refer to \cite{dblp:journals/corr/schaulqas15} for details. Each timestep, an experience tuple is perceived and added to the  replay buffer $\mathcal{D}$. Then, a minibatch of transitions is sampled from $\mathcal{D}$, on which the network is trained.
Each experience's priority is updated based on the average TD-error of the two weight vectors it was trained on.

As described in the main paper, CN can have a secondary experience replay buffer for diverse experience replay (DER). For a description of when and which samples are added to the secondary replay buffer, please refer to the main paper.

In this paper, we make use of $\varepsilon$-greedy exploration, with $\varepsilon$, the probability of performing a random action, annealed over time. For Minecart we anneal it from $1$ to $0.05$ over the first 100k steps, for the easier DST problem we anneal it to $0.01$ over 10k steps. However, CN is compatible with any sort of exploration strategy.

\subsection{Multi-Network algorithm}
The Multi-Network (MN) algorithm (Algorithm \ref{alg:multi-network}) for multi-objective deep reinforcement learning under dynamic weights handles changes in weights by gradually building an approximate partial CCS, i.e., a set of policies such that each policy performs near optimality for at least one encountered weight vector.

The algorithm starts with an empty set of policies $\Pi$. Then, for each encountered weight vector ${{\bf w}_t}$, it trains a neural network through scalarized deep Q-learning \cite{DBLP:journals/corr/MossalamARW16}. The differences with standard deep Q-learning are that the DQN's outputs are vector valued and that action selection is done by scalarizing these Q-vector-values w.r.t.\ the current weight vector ${{\bf w}_t}$ (Lines \ref{alg:MN-act} and \ref{alg:MN-opt-act} of Algorithm \ref{alg:multi-network}).

As is the case for CN, experiences are sampled from the replay buffer through prioritized sampling, with priorities being computed on the TD-error for the active weight vector.

When the active weight vector changes at time $t+1$, the policy trained (before the change) for ${\bf w}_t$ is stored if it is optimal for at least one past weight vector. To account for approximation errors, a constant $\kappa$ is subtracted from any past policy's scalarized value. Hence, when two scalarized values are within an error $\kappa$ of each other, the more recent policy is favoured. Any policy in $\Pi$ that is made redundant by the inclusion of the new policy trained on ${\bf w}_t$ is discarded.

Then, the policy with the maximal scalarized value for the new weight vector ${\bf w}_{t+1}$ is used as a starting point for its Q-network ${\bf Q}_{{\bf w}_{t+1}}$. As in \cite{DBLP:journals/corr/MossalamARW16}, we considered full re-use, where all parameters of the model Q-network are copied into the new Q-network and partial re-use, in which all but the last dense layer were copied to the new Q-network. We found that the latter performed poorly and therefore only considered full re-use in this paper.

\begin{algorithm}[!ht]
	\small
	\caption{Dynamic Multi-Network Reinforcement Learning}\label{alg:multi-network}

	\begin{algorithmic}[1]

		\STATE initialize (diverse) replay buffer $\mathcal{D}$ and unique weight history $\mathcal{W}$
		\STATE $\kappa \gets$ Improvement constant

		\STATE $\Pi \gets $ empty set of $ ( {\bf Q}_{{\bf w}}, {\bf w}, {\bf V}_{{\bf w}})$ tuples \COMMENT{With ${\bf w}$ a weight vector, ${\bf Q}_{{\bf w}}$ a policy for that weight vector (i.e., a multi-objective Q-network), ${\bf V}_{{\bf w}}$ the stateless value vector of the policy}
		\STATE ${\bf w}_0 \gets getWeightVector(0)$

		\STATE add  ${\bf w}_0$ to $\mathcal{W}$
		\STATE ${\bf Q}_{{{\bf w}_0}},{\bf Q}_{{{\bf w}_0}}^- \gets initializeFirstModel()$
		\FOR {steps $t \in \{0 ... T\}$}

		\STATE With probability $\varepsilon$ select a random action $a_t$

		\STATE Otherwise $a_t=argmax_{a\in A} {\bf Q}_{{{\bf w}_t}}(a,s) \cdot {\bf w}_t$\label{alg:MN-act}
		\STATE Execute action $a_t$ and observe ${\bf r}_t$ and ${\bf s}_{t+1}$
		\STATE Store transition $(s_t,a_t,{\bf r}_t,s_{t+1})$ in $\mathcal{D}$
		\STATE Sample minibatch of transitions from $\mathcal{D}$
		\FOR {each sampled transition $(s_j,a_j,{\bf r}_j,s_{j+1})$}

		\IF {transition is terminal}
		\STATE $ y_j = {\bf r}_j$
		\ELSE
		\STATE
		$a'_j = \argmax_{a'} {\bf Q}_{{\bf w}_t}(a',s_{j+1})\cdot{\bf w}_t$\label{alg:MN-opt-act}
		\STATE $y_j = {\bf r}_j + \gamma {\bf Q}_{{{\bf w}_t}}^-(a'_j,s_{j+1})	$
		\ENDIF

		\ENDFOR
		\STATE perform gradient descent step on
		$$\big[|y_j-{\bf Q}_{{{\bf w}_t}}(a_j,s_{j})|\big]$$
		\STATE Every $N^-$ steps; ${\bf Q}_{{{\bf w}_t}}^- = {\bf Q}_{{{\bf w}_t}}$  \COMMENT {Synchronize target network}
		\STATE anneal($\varepsilon$)

		\STATE ${\bf w}_{t+1} \gets getWeightVector(t)$

		\IF{ ${\bf w}_t \neq {\bf w}_{t+1}$}

		\IF {$\;\exists{\bf w}\in \mathcal{W} : {\bf V}_t\cdot {\bf w} > \max_{{\bf V}' \in \Pi} {\bf V}' \cdot {\bf w} - \kappa$} \label{alg:multi-network:comparison}
		\STATE add $( {\bf Q}_{{{\bf w}_{t}}}, {{\bf w}_t}, {\bf V_t})$ to $\Pi$
		\STATE remove policies made redundant by ${\bf Q}_{{{\bf w}_{t}}}$
		\ENDIF

		\STATE ${\bf Q}_{{\bf w'}},{\bf w}',{\bf V}_{\bf w'} \gets argmax_{({\bf Q}_{{\bf w'}},{\bf w}',{\bf V_{w'}}) \in \Pi} {\bf w}\cdot {\bf V_{w'}}$ \label{alg:multi-network:selectpol}  \COMMENT {pick a policy to continue learning from}
		\STATE ${\bf Q}_{{{\bf w}_{t+1}}}, {\bf Q}^-_{{{\bf w}_{t+1}}} \gets copyModel({\bf Q}_{{\bf w'}})$ \label{alg:multi-network:initmodel}  \COMMENT {Partial or full re-use}

		\STATE add  ${\bf w}_{t+1}$ to $\mathcal{W}$
		\ELSE
		\STATE ${\bf Q}_{{{\bf w}_{t+1}}},{\bf Q}_{{{\bf w}_{t+1}}}^- \gets {\bf Q}_{{{\bf w}_{t}}},{\bf Q}_{{{\bf w}_{t}}}^-$  \COMMENT{continue training same policy}

		\ENDIF
		\ENDFOR

	\end{algorithmic}

\end{algorithm}

\subsection{Diverse Experience Replay}
We now present our implementation of Diverse Experience Replay (DER, Algorithm \ref{alg:diverse-mem}).

We maintain both a first-in first-out replay buffer and a diverse replay buffer. Experiences are added to the FIFO buffer as they are observed. When the FIFO buffer is full, the oldest trace $\tau$ is removed from it and considered for memorization into the secondary diverse replay buffer.

The trace $\tau$ is only added to the secondary buffer if it increases the replay buffer diversity. To determine this, we first compute a signature for each trace up for consideration (i.e., $\tau$ and all traces already present in the diverse replay buffer $\mathcal{D}'$). Note that this signature can typically be computed in advance. Next, a diversity function $d$ computes the relative diversity of each signature w.r.t.\ all other considered signatures (Algorithm \ref{alg:diverse-mem} Line \ref{alg:der-diversity}). If $\tau$'s relative diversity is lower than the minimal relative diversity already present in the secondary buffer $\mathcal{D}'$, it is discarded. Otherwise, the trace that contributes least to the buffer's diversity is removed from $\mathcal{D}'$ to make place for $\tau$.

This process is repeated until there is enough space for $\tau$ in the diverse buffer or $\tau$ has a lower diversity than the lowest diversity trace in $\mathcal{D}'$, in which case $\tau$ is discarded and the traces that were removed during the current selection process are re-added.

For our experiments, we used a trace's return vector $\sum_{t=0}^{|\tau|} \gamma^t {\bf r}_t$ as its signature and the crowding distance \cite{nsgaii} as the diversity function.

When using DER, half of the buffer size is used by the diverse replay buffer. When sampling from DER, no distinction is made between diverse and main replay buffers.

\begin{algorithm}[!ht]
	\caption{Diverse Replay Buffer}\label{alg:diverse-mem}
	\begin{algorithmic}[1]
		\STATE   \COMMENT {With $s$ a signature function, $d$ a diversity function, $\mathcal{D}$ the main memory, $\mathcal{D}'$ the secondary memory and $e$ an experience to memorize}
		\IF {main memory $\mathcal{D}$ is full}
		\STATE extract oldest trace $\tau$ from $\mathcal{D}$
		\STATE add $e$ to $\mathcal{D}$
		\WHILE {$\mathcal{D}'$ does not have enough space for $\tau$}
		\STATE $\mathcal{F} \gets d(\{ s(\tau_i) | \tau_i \in  \mathcal{D}' \cup \{\tau\}  \} )$ \label{alg:der-diversity}

		\STATE select trace $\tau_{j}$ with lowest diversity $f_j \in \mathcal{F}$
		\IF {$\tau_{j} \neq \tau$}
		\STATE remove $\tau_{j}$ from $\mathcal{D}'$
		\ELSE
		\STATE Discard $\tau$
		\STATE undo deletions
	 \STATE \textbf{return}
		\ENDIF
		\ENDWHILE
		\STATE add $\tau$ to $\mathcal{D}'$
		\ENDIF
	\end{algorithmic}
\end{algorithm}

\section{Implementation details}
We now present our implementation details. More specifically, we give a table of hyperparameters (Table \ref{table:hyperparameters}) and a full description of the network architecture.
\begin{table}
	\caption{Hyperparameters}\label{table:hyperparameters}
	\centering
	\begin{tabular}{|c|c|}
		\hline
		\multicolumn{2}{|c|}{General parameters (Minecart)} \\
		\hline
		Exploration rate  & $1 \rightarrow 0.05$ over 100k steps \\
		Buffer size       & 100.000                              \\
		Frame skip        & 4                                    \\
		Discount factor   & 0.98                                 \\
		\hline
		\multicolumn{2}{|c|}{General parameters (DST)} \\
		\hline
		Exploration rate  & $0.1 \rightarrow 0.01$ over 10k steps  \\
		Buffer size       & 10.000                               \\
		Frame skip        & 1                                    \\
		Discount factor   & 0.95                                 \\
		\hline
		\multicolumn{2}{|c|}{Optimization parameters} \\
		\hline
		Batch size        & 64 (Minecart), 16 (DST)              \\
		Optimizer         & SGD                                  \\
		Learning rate     & 0.02                                 \\
		Momentum          & 0.90                                    \\
		Nesterov Momentum & true                                 \\
		$N^-$             & 150                                  \\

		\hline
		\multicolumn{2}{|c|}{Prioritized sampling parameters} \\
		\hline
		$\varepsilon$     & 0.01                                 \\
		$\alpha$          & 2.0                                  \\
		\hline
	\end{tabular}
\end{table}

\subsection{Network Architecture}
\label{architecture}
\begin{figure}[t]
	\center
	\includegraphics[width=.45\textwidth]{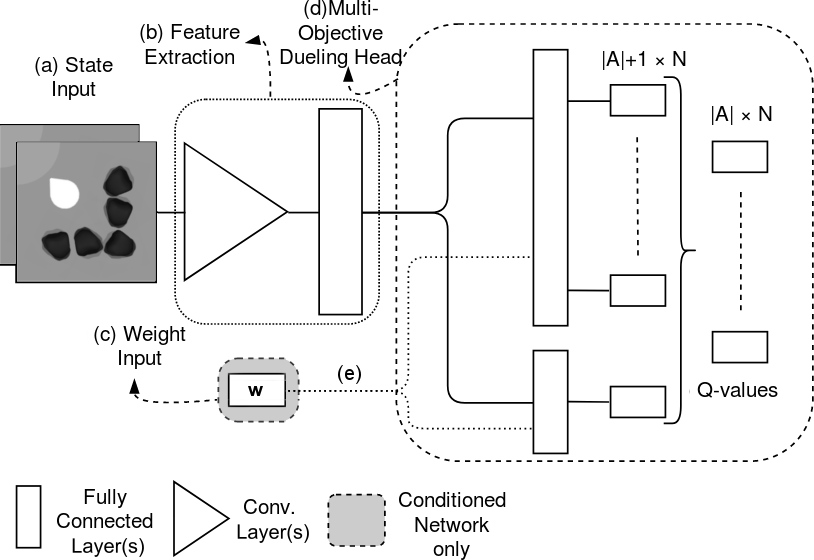}
	\caption{Features are extracted from the raw input by convolutional layers followed by a fully connected layer. The extracted features (output of (b)) are fed into a Multi-Objective Dueling DQN head (d). The conditioned architecture feeds a weight input (c) into the Q-value head (link (e)).}
	\label{fig:architecture}
\end{figure}
Figure \ref{fig:architecture} gives a schematic representation of the architecture we used in our experiments.

Our network contains more dense layers than single-objective Dueling DQN, we justify this by the need to output multi-objective Q-values and to either (1) output precise Q-values (in the case of MN knowing one action is better than another is not sufficient), or (2) learn multiple weight-conditioned policies (in the case of CN). 

The input to the network consists of two 48x48 frames (scaled down from the original 480x480 dimensions). The first convolution layer consists of 32 6x6 filters with stride 2. The second convolution layer consists of 48 5x5 filters with stride 2. Each convolution is followed by a maxpooling layer. A dense layer of 512 units is then connected to each temporal dimension of the convolution. 

Following a multi-objective generalization of the \emph{Dueling DQN} architecture \cite{dblp:journals/corr/wangfl15}, this layer's outputs are then fed into the advantage and value streams, consisting of dense layers of size 512. The advantage stream is then fed into a layer of $|A| \times N$ dense units, while the value stream is fed into a dense layer of $N$ units. These $|A|+1 \times N$ outputs are then combined into $|A| \times N$ outputs by a multi-objective generalization of Dueling DQN's module.

\begin{multline}
	{\bf Q}(s,a;\theta,\alpha,\beta) = {\bf V}(s;\theta,\beta) + \\
	\Big( {\bf A}(s,a;\theta,\alpha) - \frac{1}{|\mathcal{A}|}\sum_{a'}{\bf A}(s,a';\theta,\alpha)  \Big)
\end{multline}
$\alpha$ and $\beta$ denote the parameters of the advantage stream and of the value stream and $\theta$ denotes the parameters of all preceding layers.
For the Conditioned Network, an additional parameter ${\bf w}$ is added to each function (corresponding to the weight input, link (e) in Figure \ref{fig:architecture});
\begin{multline}
	{\bf Q}(s,a,{\bf w};\theta,\alpha,\beta) = {\bf V}(s,{\bf w};\theta,\beta) +\\
	\Big( {\bf A}(s,a,{\bf w};\theta,\alpha) - \frac{1}{|\mathcal{A}|}\sum_{a'}{\bf A}(s,a',{\bf w};\theta,\alpha)  \Big)
\end{multline}

The hyperparameters used for optimization are given in Table \ref{table:hyperparameters}.

\section{Test Problems}
In this section, we present the test problems used in our experimental evaluation in greater detail.

\subsection{Minecart Problem}

The Minecart problem models the challenges of resource collection, has a continuous state space, stochastic transitions and delayed rewards. %

\begin{figure}
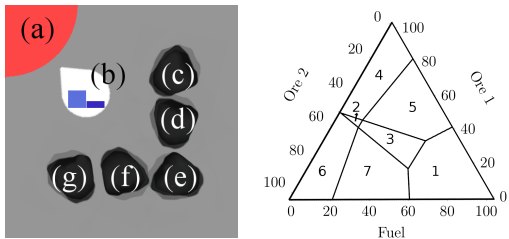

	\center
	\includegraphics[width=.18\textwidth]{img/testsetting}~~~
	\includegraphics[width=.2\textwidth]{img/weightspace}
	\caption{(left) Instance of the Minecart environment with 5 mines ((c) to (g)) containing varying amounts of 2 ores. The 2 bars on the minecart (b) indicate how much of each ore is present in the cart. Ores are sold on the base (a). (right) Weight vectors in the same region share the same optimal policy. Axes are the relative importance in \% of each objective. We distinguish (1) collecting no resources if the fuel cost is too high, (6,7) privileging ore 2, (4,5) privileging ore 1, and (2,3) privileging the quick collection of either ore. Differences between each pair lies in the higher fuel cost, in which case it is optimal to accelerate less.}
	\label{fig:minecart}
\end{figure}
The Minecart environment consists of a rectangular image, depicting a base, mines and the minecart controlled by the agent.
A typical frame of the Minecart environment is given in Figure \ref{fig:minecart} (left). Each episode starts with the agent on top of the base. Through the \textit{accelerate}, %
\textit{brake}, %
\textit{turn left}, %
\textit{turn right}, %
\textit{mine}, or \textit{do nothing} (useful to preserve momentum) actions,
the agent should reach a mine, collect resources from it and return to the base to sell the collected resources.

The reward vectors are N-dimensional: ${\bf r}=(r_1,...,r_N)$. The first $N-1$ elements correspond to the amount of each of the $N-1$ resources the agent gathered, the last element is the consumed fuel. Particular challenges of this environment are the sparsity of the first $N-1$ components of the reward vector, %
as well as the delay between actions (e.g., mining) and resulting reward. %
The resources an agent collects by mining are generated from the mine's random distribution, resulting in a stochastic transition function. All other actions result in deterministic transitions.
The weight vector ${\bf w}$ expresses the relative importance of each objective, i.e., the price per resource. %

The underlying state consists of the minecart's position, its velocity and its content, the position of the mines and their respective ore distribution. While the implementation makes these available for non-deep MORL research, these properties were not used in our experiments. In the deep setting the agent should learn to extract them from the visual representation of the state. %

Figure \ref{fig:minecart} shows the visual cues the agent should exploit to extract appropriate features of the state. First and most obviously, the position of the mines (black) and the home position (area top left). Second, indicators about the minecart's content are represented by vertical bars on the cart, one for each ore type. Each bar is the size of the cart's capacity. %
When the cart has reached its maximal capacity $C$, and mining will have no effect on the cart's content but still incur the normal mining penalty $p_m$ in terms of fuel consumption. At that point the agent should return back to its home position. Additionally, the minecart's orientation is given by the cone's direction. Accelerating incurs a penalty of $p_a$ in terms of fuel consumption. In addition, every time-step the agent receives a penalty in the fuel objective $p_i$ representing the cost of keeping the engine running.

The default configuration of the minecart environment we used in our experiments is given in Table \ref{table:mineconfig}. The setting contains 5 mines, with distribution means for ores 1 and 2 given in Table \ref{table:distributions}, and a standard deviation fixed at $\sigma=0.05$.

\paragraph{Optimal Policies}
For $\gamma=0.98$ used in our experiments, this configuration divides the weight-space into 7 regions according to their optimal policies as shown in Figure \ref{fig:minecart}. The 7 policies are;
\begin{enumerate}
	\item do not collect any resources
	\item go to mine (e) quickly and mine until full,
	\item go to mine (e) slowly and mine until full,
	\item go to mine (c) rapidly and mine until full,
	\item go to mine (c) slowly and mine until full,
	\item go to mine (g) quickly and mine until full,
	\item go to mine (g) slowly and mine until full,
\end{enumerate}

\begin{table}
	\caption{Minecart configuration}\label{table:mineconfig}
	\centering
	\begin{tabular}{|c|c|}
		\hline
		\multicolumn{2}{|c|}{General Minecart Configuration} \\
		\hline
		Cart capacity           & 1.5        \\
		Acceleration            & 0.0075     \\
		Ores                    & 2          \\
		Rotation angle          & 10 degrees \\
		\hline
		\multicolumn{2}{|c|}{Fuel component rewards} \\
		\hline
		Idle cost $p_i$         & -0.005     \\
		Mining cost $p_m$       & -0.05      \\
		Acceleration cost $p_a$ & -0.025     \\
		\hline
	\end{tabular}
\end{table}

\begin{table}
	\caption{Ore distribution per mine, if either ore is more valuable, mining from (d) to (f) results in wasted capacity on the less valuable ore. Hence, while the average content collected from these mines is higher, they are not always optimal because of the limited cart capacity.}\label{table:distributions}
	\begin{center}

		\begin{tabular}{ c | c | c | c | c | c }
			Mine          & (c) & (d) & (e) & (f) & (g) \\
			\hline
			$\mu_{ore_1}$ & 0.2 & 0.15 & 0.2 & 0.1 & 0.  \\
			$\mu_{ore_2}$ & 0.  & 0.1 & 0.2 & 0.15 & 0.2
		\end{tabular}

	\end{center}

\end{table}

\subsection{Deep Sea Treasure}
\begin{figure}[!ht]
	\center
	\includegraphics[width=.33\textwidth]{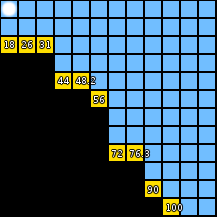}
	\caption{DST map, yellow squares indicate treasures and their value, the agent is marked by a white circle. Black areas are the ocean floor, blue areas are the ocean.}
	\label{fig:dst-map}
\end{figure}

In the Deep Sea Treasure (DST) problem \cite{Vamplew2011}, a submarine must dive to collect a treasure. The further the treasure is, the higher its value. The agent can move \emph{left}, \emph{right}, \emph{up}, and \emph{down} which will move him to the corresponding neighboring cell unless that cell is outside of the map or a sea bottom cell (black cells). The reward signal an agent perceives consists of a treasure component and a time component. The submarine collects a penalty of $-1$ for its second objective at every step. When it reaches a treasure, the treasure's value is collected as a reward for the first objective, and the episode ends. As the original DST problem has only two policies in its convex coverage set, we used a modified version of the DST map -- given in Figure \ref{fig:dst-map} -- in our experiments.

\begin{figure}[!ht]
	\center
	\includegraphics[width=.33\textwidth]{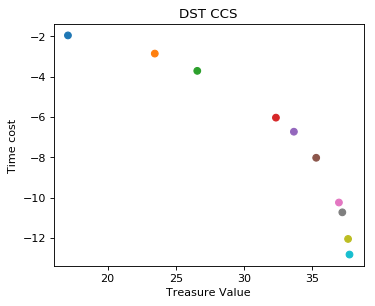}
	\caption{Convex Coverage Set for the given DST map and a discount factor of $\gamma=0.95$.}
	\label{fig:dst-ccs}
\end{figure}

\begin{figure}[!ht]
	\center
	\includegraphics[width=.33\textwidth]{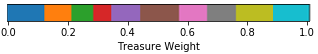}
	\caption{Optimal policy colormap, each color corresponds to one of the optimal policies in Figure \ref{fig:dst-ccs}. Time cost weight is $1 -$ treasure weight.}
	\label{fig:dst-division}
\end{figure}

 This map was designed such that, for a discount factor of $0.95$, each treasure is the goal of an optimal policy in the CCS (Figure \ref{fig:dst-ccs}). And in addition, each policy in the CCS has approximately the same proportion of weights for which it is optimal (${\sim}10\%$ of weight vectors for each policy, Figure \ref{fig:dst-division}).
The full results obtained for the image version DST are given in Table \ref{table:results-dst}.


\section{Additional Results}
In this section we give the complete results table for DST (Table \ref{table:results-dst}), and we experimentally compare selective experience replay \cite{DBLP:journals/corr/abs-1802-10269} and Exploration-based selection \cite{de2018experience} to DER on the Minecart problem.
We also provide results for a naive baseline (NAIVE) suggest by \cite{6918520}. 
\begin{table*}[!ht]
	
	\small
	\caption{Average episodic regret ($\Delta$) and improvement over MO with Std.\ ER baseline ($>$) for both weight change scenarios (lower is better) for DST. %
		We distinguish between overall performance, and performance over the last 25k steps}\label{table:results-dst}
	\centering
	\begin{tabular}{|c|c|l|l|l|l|l|l|l|l|}

		\cline{3-10}

		\multicolumn{2}{c|}{}& \multicolumn{4}{|c|}{Overall} & \multicolumn{4}{|c|}{Last 25k steps}\\

		\cline{3-10}

		\multicolumn{2}{c|}{}& \multicolumn{2}{|c|}{Standard ER}       & \multicolumn{2}{|c|}{DER}               & \multicolumn{2}{|c|}{Standard ER}       & \multicolumn{2}{|c|}{DER}        \\

		\cline{2-10}

		\multicolumn{1}{c|}{} & Algorithm & $\Delta$ & $>$        & $\Delta$       & $>$                & $\Delta$ & $>$        & $\Delta$       & $>$                 \\

		\hline

		& NAIVE & 0.061 & +64.86\% & 0.06 & +62.16\% & 0.064 & +137.04\% & 0.084 & +211.11\% \\

		& MO & 0.037 & -0.0\% & 0.031 & -16.22\% & 0.027 & -0.0\% & 0.022 & -18.52\% \\
		\cline{2-10}
Sparse & MN & 0.031 & -16.22\% & 0.03 & -18.92\% & 0.02 & -25.93\% & 0.019 & -29.63\% \\
\cline{2-10}
Weight & CN & 0.024 & -35.14\% & \textbf{0.021} & \textbf{-43.24\%} & 0.012 & -55.56\% & \textbf{0.009} & \textbf{-66.67\%} \\
Changes & CN-UVFA & 0.025 & -32.43\% & 0.023 & -37.84\% & 0.015 & -44.44\% & \textbf{0.009} & \textbf{-66.67\%} \\
& CN-ACTIVE & 0.032 & -13.51\% & 0.028 & -24.32\% & 0.021 & -22.22\% & 0.016 & -40.74\% \\
\cline{2-10}
& UVFA & 0.034 & -8.11\% & 0.03 & -18.92\% & 0.023 & -14.81\% & 0.017 & -37.04\% \\

				\hline
				& NAIVE & 0.093 & +97.87\% & 0.095 & +102.13\% & 0.1 & +122.22\% & 0.114 & +153.33\% \\
& MO & 0.047 & -0.0\% & 0.052 & +10.64\% & 0.045 & -0.0\% & 0.05 & +11.11\% \\
\cline{2-10}
Regular & MN & 0.113 & +140.43\% & 0.111 & +136.17\% & 0.126 & +180.0\% & 0.104 & +131.11\% \\
\cline{2-10}
Weight & CN & 0.029 & -38.3\% & \textbf{0.025} & \textbf{-46.81\%} & 0.02 & -55.56\% & \textbf{0.014} & \textbf{-68.89\%} \\
Changes & CN-UVFA & 0.029 & -38.3\% & 0.028 & -40.43\% & 0.018 & -60.0\% & 0.017 & -62.22\% \\
& CN-ACTIVE & 0.04 & -14.89\% & 0.042 & -10.64\% & 0.03 & -33.33\% & 0.032 & -28.89\% \\
\cline{2-10}
& UVFA & 0.057 & +21.28\% & 0.051 & +8.51\% & 0.053 & +17.78\% & 0.046 & +2.22\% \\			
		\hline
	\end{tabular}
	
\end{table*}

\subsection{Naive algorithm}
The naive algorithm suggested by \cite{6918520} learns optimal Q-values for each objective then selects actions by scalarizing these multiple single-objective Q-values can learn to perform well for edge weight vectors (i.e., weight vectors for which one objective is much more important than the others). However, when a trade-off is required between objectives it would be unable to perform optimally.

\subsection{Exploration-based Selection}
\citeauthor{de2018experience} (\citeyear{de2018experience}) propose an alternative to FIFO memorization based on how exploratory a transition's action $a_t$ is. The distance between the Q-value of the optimal action $a_t^*$ in state $s_t$ and the taken action $a_t$ are used as a diversity metric. Hence actions that differ strongly from the optimal action are more likely to be preserved in the replay buffer. The main obstacle to this approach working in this setting is that the exploratory nature of an action is likely to be dependent on the weight vector. An action that would be exploratory for a weight vector $\bf w$ could be optimal for another weight vector $\bf w'$. We found that while this metric can be useful for a single weight vector, it proves unreliable when used across different weight vectors. 
In addition, we identified the long-term dependence there can be between experiences. In complex problems, a reward is typically the result of a long sequence of actions. Hence, while exploration-based selection might permanently store an interesting experience, it is unlikely to store all the experiences leading to that experience. In contrast, our approach handles trajectories as atomic units. Hence, if a rewarding experience is stored, the actions leading to that reward will be stored too.

\subsection{Selective Experience Replay}
Selective experience replay \cite{DBLP:journals/corr/abs-1802-10269} was recently proposed to prevent catastrophic forgetting in single-objective multi-task lifelong learning. In this setting, an agent must learn to perform well on a sequence of tasks and maintain that performance while learning new tasks. As a result the replay buffer can be biased towards the most recent task. From there, a parallel can be drawn with the multiple policies that need to be learned for different ${\bf w}_t$ in our setting, and the resulting bias. While some of the challenges of both settings are comparable, we found that selective experience replay performs poorly on our dynamic weights problem. We hypothesize that this is due to two major differences in our approach.
First, the transition-based selection presents the same problem we observe for Exploration-based Selection (see above).
Second, their best working variant of selective experience replay, called \textit{distribution matching} does not promote diverse experiences, instead it attempts to match the distribution of experiences across all tasks. If this distribution is not diverse, rare interesting experiences obtained through random exploration are likely to be overridden by more common experiences.

\subsection{Results}
\begin{figure*}[t]
	\centering
	\begin{subfigure}[t]{0.48\textwidth} 
		\includegraphics[width=\textwidth, trim=0.2mm 0mm 0mm 0mm, clip=true]{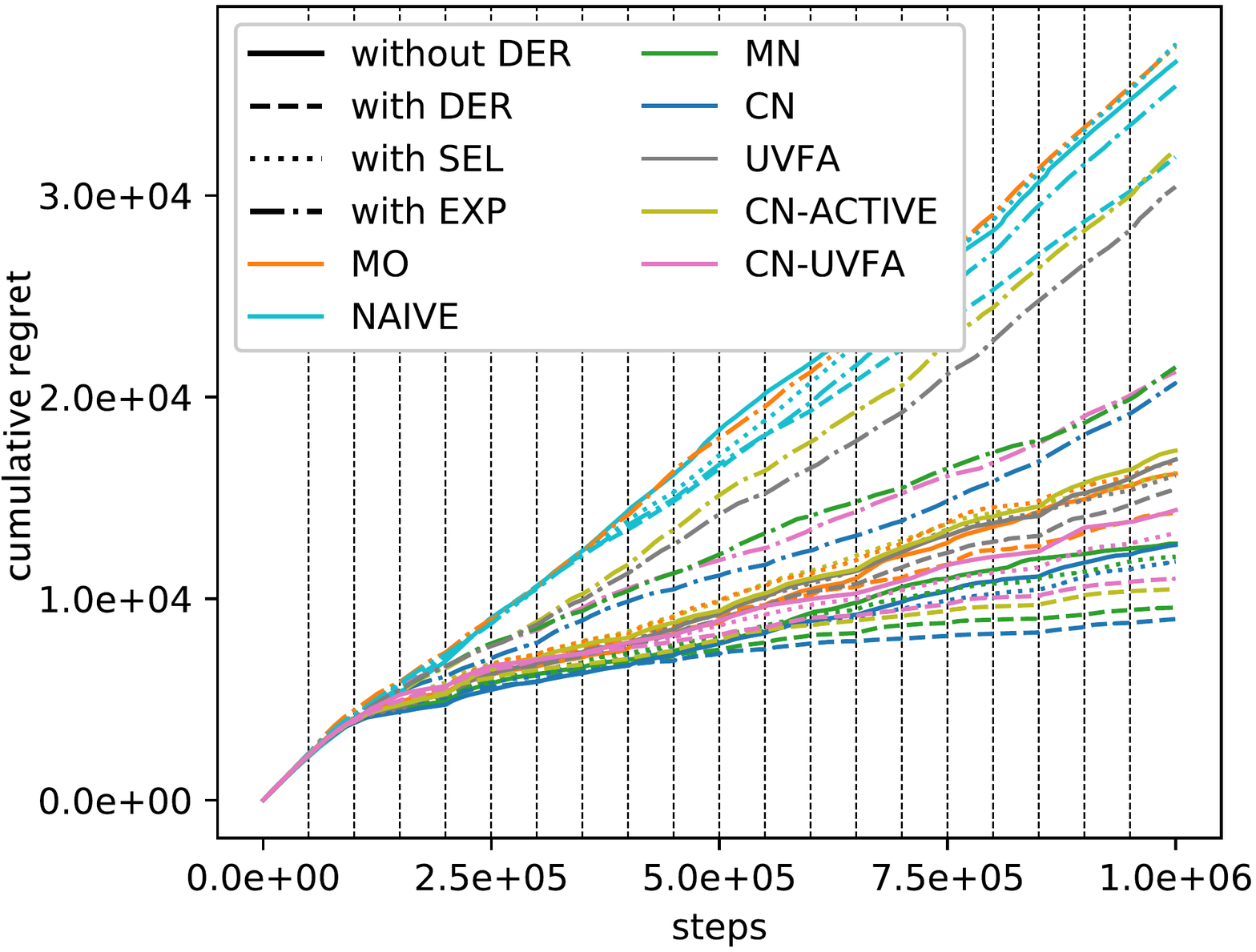}
	\end{subfigure}%
	\begin{subfigure}[t]{0.48\textwidth}
		\includegraphics[width=\textwidth]{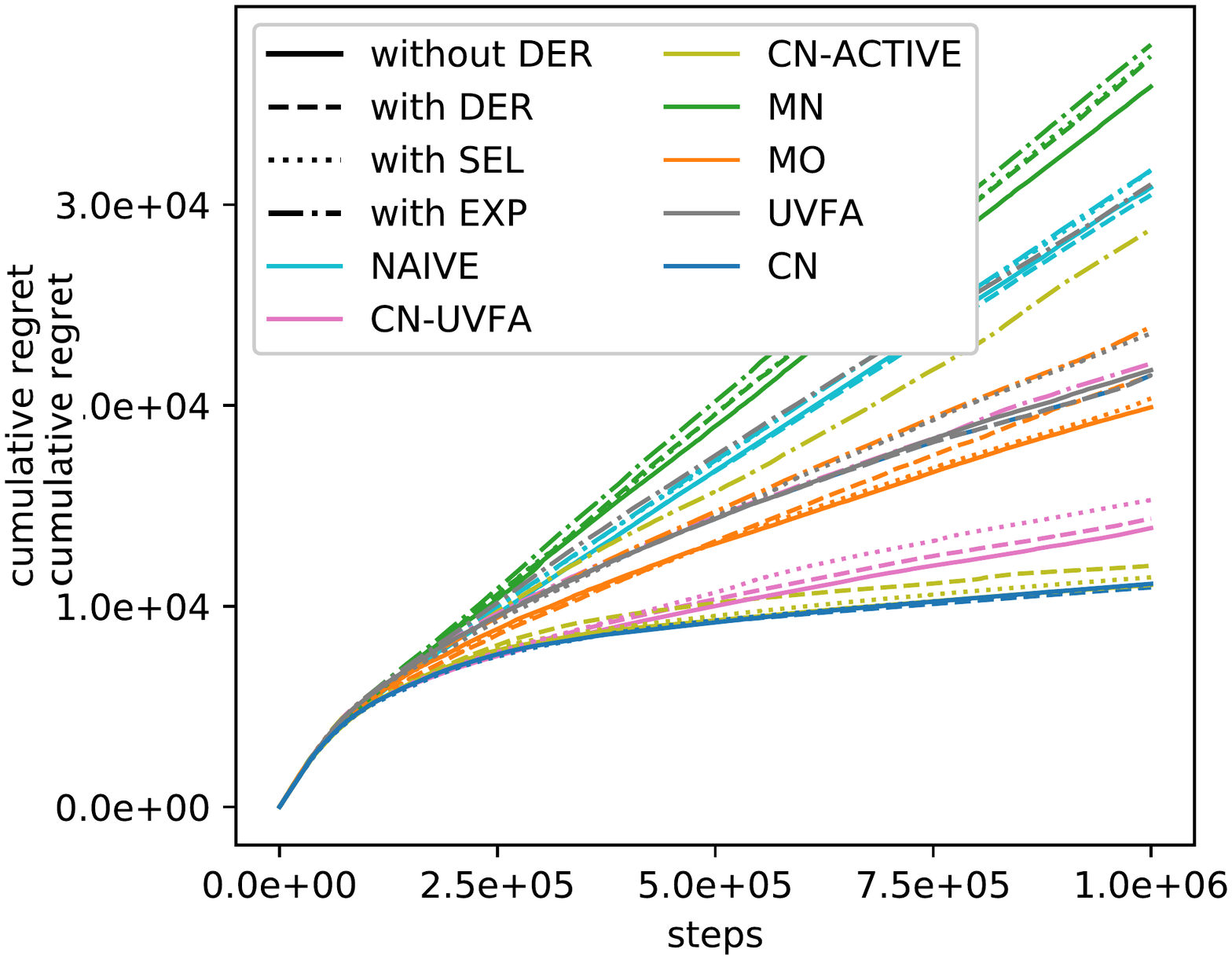}
	\end{subfigure}
	\caption{\textbf{Left:} Cumulative regret for the Minecart problem when weights change every 50k steps (vertical lines), MN+DER and OnUVFA+DER overlap each other, CN+SEL and MN+SEL overlap each other. \textbf{Right:} Cumulative regret for the Minecart problem when weights change over the span of 10 episodes, CN, CN+DER, CN+SEL and CNC overlap to form the lowest curve.} \label{fig:mc-plots-sel}
\end{figure*}

These factors contribute to the poor performance of exploration-based selection and selective experience replay (which we label respectively as EXP and SEL in Figure \ref{fig:mc-plots-sel}  and Tables \ref{table:sel-result-table-sparse-overall},\ref{table:sel-result-table-sparse-final},\ref{table:sel-result-table-reg-overall} and \ref{table:sel-result-table-reg-final}). For all algorithms in the sparse weight change scenario, selective experience replay performs worse than DER. However, we found that SEL generally improved performance over standard experience replay. In contrast, EXP has a consistently damaging effect on performance. 
As for DER, we find that the influence of SEL on the regular weight change scenario is insignificant. EXP however still has a significant negative impact on performance.
Regardless of the weight change scenario or experience replay type, the naive algorithm fails to learn any kind of tradeoff and as a result it performs poorly across our experiments.

\begin{table*}[!h]
	\caption{Overall average episodic regret ($\Delta$) and improvement over MO with Std.\ ER baseline ($>$) for the \emph{sparse weight change} scenario (lower is better) for \emph{Minecart}. %
	}\label{table:sel-result-table-sparse-overall}
	\small
	
	\centering
	\begin{tabular}{|c|l|l|l|l|l|l|l|l|}

		\cline{2-9}

		\multicolumn{1}{c|}{}& \multicolumn{8}{|c|}{Overall} \\

		\cline{2-9}

		\multicolumn{1}{c|}{}& \multicolumn{2}{|c|}{Standard ER}       & \multicolumn{2}{|c|}{DER}   & \multicolumn{2}{|c|}{SEL}  & \multicolumn{2}{|c|}{EXP}                 \\

		\cline{1-9}

		Algorithm & $\Delta$ & $>$     & $\Delta$ & $>$     & $\Delta$ & $>$     & $\Delta$ & $>$       \\

		\cline{1-9}
		NAIVE & 0.732 & +125.93\% & 0.638 & +96.91\% & 0.75 & +131.48\% & 0.709 & +118.83\% \\
MO & 0.324 & $-$ & 0.285 & -12.04\% & 0.336 & +3.7\% & 0.748 & +130.86\% \\
\hline
MN & 0.255 & -21.3\% & 0.191 & -41.05\% & 0.242 & -25.31\% & 0.43 & +32.72\% \\
\hline
CN & 0.253 & -21.91\% & \textbf{0.18} & \textbf{-44.44\%} & 0.237 & -26.85\% & 0.414 & +27.78\% \\
CN-UVFA & 0.288 & -11.11\% & 0.22 & -32.1\% & 0.265 & -18.21\% & 0.425 & +31.17\% \\
CN-ACTIVE & 0.347 & +7.1\% & 0.21 & -35.19\% & 0.325 & +0.31\% & 0.645 & +99.07\% \\
\hline
UVFA & 0.338 & +4.32\% & 0.308 & -4.94\% & 0.322 & -0.62\% & 0.609 & +87.96\%  \\
		
\hline

	\end{tabular}

\end{table*}

\begin{table*}[!h]
	\caption{Average episodic regret ($\Delta$) and improvement over MO with Std.\ ER baseline ($>$) for the \emph{sparse weight change} scenario (lower is better) for \emph{Minecart} over the last 250k steps. %
		}\label{table:sel-result-table-sparse-final}
	\small
	\centering
	\begin{tabular}{|c|l|l|l|l|l|l|l|l|}

		\cline{2-9}

		\multicolumn{1}{c|}{}& \multicolumn{8}{|c|}{Last 250k steps} \\

		\cline{2-9}

		\multicolumn{1}{c|}{}& \multicolumn{2}{|c|}{Standard ER}       & \multicolumn{2}{|c|}{DER}   & \multicolumn{2}{|c|}{SEL}  & \multicolumn{2}{|c|}{EXP}                 \\

		\cline{1-9}

		Algorithm & $\Delta$ & $>$     & $\Delta$ & $>$     & $\Delta$ & $>$     & $\Delta$ & $>$       \\

		\cline{1-9}
		NAIVE & 0.791 & +187.64\% & 0.651 & +136.73\% & 0.851 & +209.45\% & 0.802 & +191.64\% \\
MO & 0.275 & $-$ & 0.207 & -24.73\% & 0.241 & -12.36\% & 0.818 & +197.45\% \\
\hline
MN & 0.139 & -49.45\% & \textbf{0.063} & \textbf{-77.09\%} & 0.133 & -51.64\% & 0.403 & +46.55\% \\
\hline
CN & 0.184 & -33.09\% & 0.068 & -75.27\% & 0.155 & -43.64\% & 0.467 & +69.82\% \\
CN-UVFA & 0.218 & -20.73\% & 0.102 & -62.91\% & 0.187 & -32.0\% & 0.414 & +50.55\% \\
CN-ACTIVE & 0.316 & +14.91\% & 0.088 & -68.0\% & 0.202 & -26.55\% & 0.754 & +174.18\% \\
\hline
UVFA & 0.302 & +9.82\% & 0.253 & -8.0\% & 0.213 & -22.55\% & 0.743 & +170.18\%  \\
\hline

	\end{tabular}
	
\end{table*}

\begin{table*}[!h]
	\caption{Overall Average episodic regret ($\Delta$) and improvement over MO with Std.\ ER baseline ($>$) for the \emph{regular weight change} scenario (lower is better) for Minecart. %
	}\label{table:sel-result-table-reg-overall}
	\small
	\centering
	\begin{tabular}{|c|l|l|l|l|l|l|l|l|}

		\cline{2-9}

		\multicolumn{1}{c|}{}& \multicolumn{8}{|c|}{Overall} \\

		\cline{2-9}

		\multicolumn{1}{c|}{}& \multicolumn{2}{|c|}{Standard ER}       & \multicolumn{2}{|c|}{DER}   & \multicolumn{2}{|c|}{SEL}  & \multicolumn{2}{|c|}{EXP}                 \\

		\cline{1-9}

		Algorithm & $\Delta$ & $>$     & $\Delta$ & $>$     & $\Delta$ & $>$     & $\Delta$ & $>$       \\

		\cline{1-9}
		NAIVE & 0.617 & +55.03\% & 0.61 & +53.27\% & 0.634 & +59.3\% & 0.635 & +59.55\% \\
MO & 0.398 & $-$ & 0.43 & +8.04\% & 0.407 & +2.26\% & 0.478 & +20.1\% \\
\hline
MN & 0.718 & +80.4\% & 0.746 & +87.44\% & 0.748 & +87.94\% & 0.76 & +90.95\% \\
\hline
CN & 0.222 & -44.22\% & \textbf{0.219} & \textbf{-44.97\%} & 0.222 & -44.22\% & 0.43 & +8.04\% \\
CN-UVFA & 0.278 & -30.15\% & 0.287 & -27.89\% & 0.306 & -23.12\% & 0.442 & +11.06\% \\
CN-ACTIVE & 0.221 & -44.47\% & 0.24 & -39.7\% & 0.229 & -42.46\% & 0.576 & +44.72\% \\
\hline
UVFA & 0.435 & +9.3\% & 0.43 & +8.04\% & 0.472 & +18.59\% & 0.62 & +55.78\% \\
\hline

	\end{tabular}

\end{table*}

\begin{table*}[!h]
	\caption{Average episodic regret ($\Delta$) and improvement over MO with Std.\ ER baseline ($>$) for the \emph{regular weight change} scenario (lower is better) for Minecart over the last 250k steps. %
	}\label{table:sel-result-table-reg-final}
	\small
	\centering
	\begin{tabular}{|c|l|l|l|l|l|l|l|l|}

		\cline{2-9}

		\multicolumn{1}{c|}{}& \multicolumn{8}{|c|}{Last 250k steps} \\

		\cline{2-9}

		\multicolumn{1}{c|}{}& \multicolumn{2}{|c|}{Standard ER}       & \multicolumn{2}{|c|}{DER}   & \multicolumn{2}{|c|}{SEL}  & \multicolumn{2}{|c|}{EXP}                 \\

		\cline{1-9}

		Algorithm & $\Delta$ & $>$     & $\Delta$ & $>$     & $\Delta$ & $>$     & $\Delta$ & $>$       \\

		\cline{1-9}
		NAIVE & 0.56 & +117.05\% & 0.551 & +113.57\% & 0.588 & +127.91\% & 0.581 & +125.19\% \\
MO & 0.258 & $-$ & 0.319 & +23.64\% & 0.251 & -2.71\% & 0.36 & +39.53\% \\
\hline
MN & 0.67 & +159.69\% & 0.709 & +174.81\% & 0.72 & +179.07\% & 0.712 & +175.97\% \\
\hline
CN & 0.069 & -73.26\% & \textbf{0.064} & \textbf{-75.19\%} & 0.066 & -74.42\% & 0.267 & +3.49\% \\
CN-UVFA & 0.149 & -42.25\% & 0.149 & -42.25\% & 0.165 & -36.05\% & 0.299 & +15.89\% \\
CN-ACTIVE & 0.065 & -74.81\% & 0.071 & -72.48\% & 0.069 & -73.26\% & 0.56 & +117.05\% \\
\hline
UVFA & 0.273 & +5.81\% & 0.267 & +3.49\% & 0.346 & +34.11\% & 0.538 & +108.53\% \\
		\hline

	\end{tabular}
	
\end{table*}

\end{document}